\documentclass[10pt,twocolumn,letterpaper]{article}

\usepackage[pagenumbers]{cvpr} % To force page numbers, e.g. for an arXiv version

\usepackage{wrapfig}
\usepackage{overpic}
\usepackage{enumitem} %< control spacing in itemize/enumerate/...
\usepackage{overpic} %< add raw math symbols to figures
\usepackage{color}
\usepackage{comment}
\usepackage{booktabs} % For formal tables
\usepackage[accsupp]{axessibility} % Improves PDF readability for those with visual impairments.

\usepackage[ruled,linesnumbered]{algorithm2e}
\usepackage{amsfonts}
\usepackage{threeparttable}
\usepackage{tabularx}
\usepackage{mathtools}

\usepackage{caption}
\usepackage{subcaption}
\usepackage{multirow}
\usepackage{algpseudocode}
\usepackage{soul}
\usepackage{multirow}
\usepackage{arydshln} % for dashed line
\usepackage[normalem]{ulem}
\usepackage[most]{tcolorbox}
\usepackage{textcomp}
\usepackage[hang,flushmargin]{footmisc}

\usepackage{amsmath,amsfonts,bm,mathtools}

\def\eqref#1{equation~\ref{#1}}

\def\1{\bm{1}}

\newcommand{\bI}{\mathbf{I}}

\newcommand{\bx}{\mathbf{x}}
\newcommand{\by}{\mathbf{y}}

\newcommand{\bepsilon}{{\boldsymbol{\epsilon}}}

\DeclareMathAlphabet{\mathsfit}{\encodingdefault}{\sfdefault}{m}{sl}
\SetMathAlphabet{\mathsfit}{bold}{\encodingdefault}{\sfdefault}{bx}{n}

\newcommand{\Hbar}{\text{\kern-0.0em\resizebox{!}{1.5ex}{H}\kern-0.8em\raisebox{0.6ex}{—}}}

\DeclareRobustCommand{\Colon}{{%
  \ooalign{%
    \hidewidth\raisebox{0.2ex}{/}\kern0.1em\hidewidth\cr
    C\cr
    \hidewidth\kern0.1em\raisebox{0.2ex}{/}\hidewidth\cr
  }%
}}

\definecolor{turquoise}{cmyk}{0.65,0,0.1,0.3}
\definecolor{purple}{rgb}{0.65,0,0.65}
\definecolor{dark_green}{rgb}{0, 0.5, 0}
\definecolor{orange}{rgb}{0.8, 0.6, 0.2}
\definecolor{red}{rgb}{0.9, 0.1, 0.1}
\definecolor{darkred}{rgb}{0.6, 0.1, 0.05}
\definecolor{blueish}{rgb}{0.0, 0.3, .6}
\definecolor{light_gray}{rgb}{0.7, 0.7, .7}
\definecolor{pink}{rgb}{1, 0, 1}
\definecolor{greyblue}{rgb}{0.25, 0.25, 1}

\newcommand{\rev}[1]{#1}

\newcommand{\methodname}{VecFusion\xspace}
\newcommand{\bff}{\mathbf{f}}
\newcommand{\bg}{\mathbf{g}}

\usepackage{blindtext}

\ifpdf

\else
\fi

\ifpdf

\else
\fi

\ifpdf

\else
\fi

\usepackage{color}
\definecolor{cyan}{cmyk}{1,0,0,0}
\definecolor{darkgreen}{rgb}{0,0.5,0}
\definecolor{orange}{rgb}{1,0.5,0}
\definecolor{magenta}{cmyk}{0,1,0,0}
\definecolor{darkyellow}{cmyk}{0,0,0.75,0}
\definecolor{gray}{rgb}{0.8,0.8,0.8}
\definecolor{good}{rgb}{0.75,0.9,0.75}
\definecolor{decent}{rgb}{0.9,0.93,0.75}
\definecolor{bad}{rgb}{0.9,0.75,0.75}
\definecolor{na}{rgb}{0.8,0.8,0.8}

\definecolor{blueteaser}{HTML}{1a85ff}
\definecolor{redteaser}{HTML}{d41159}

\definecolor{fontgreen}{RGB}{0,190,0}
\definecolor{glyphblue}{RGB}{0,0,220}

\newcommand\blfootnote[1]{%
  \begingroup
  \renewcommand\thefootnote{}\footnote{#1}%
  \endgroup
}

\usepackage{graphicx}
\usepackage{amsmath}
\usepackage{amssymb}
\usepackage{booktabs}
\usepackage{afterpage}

\definecolor{cvprblue}{rgb}{0.21,0.49,0.74}
\usepackage[pagebackref,breaklinks,colorlinks,citecolor=cvprblue]{hyperref}

\usepackage[capitalize]{cleveref}
\crefname{section}{Sec.}{Secs.}
\Crefname{section}{Section}{Sections}
\Crefname{table}{Table}{Tables}
\crefname{table}{Tab.}{Tabs.}

\begin{document}

\title{\methodname : Vector Font Generation with Diffusion}

\author{
Vikas Thamizharasan*$^{1,2}$~~~ 
Difan Liu*$^2$~~~
Shantanu Agarwal$^1$~~~
Matthew Fisher$^{2}$~~~\\
Michaël Gharbi$^2$~~~
Oliver Wang$^3$~~~
Alec Jacobson$^{2,4}$~~~
Evangelos Kalogerakis$^1$~~~
\vspace{0.3cm} \\
{\small
University of Massachusetts Amherst$^1$~~~
Adobe Research$^2$~~~
Google Research$^3$~~~
University of Toronto$^4$
}
%\vspace{0.2cm}
}
% \date{\vspace{-3ex}}

\twocolumn[{%
 \renewcommand\twocolumn[1][]{#1}%
 \maketitle
 \vspace{-5mm}
 \centering
\includegraphics[width=0.9\textwidth]{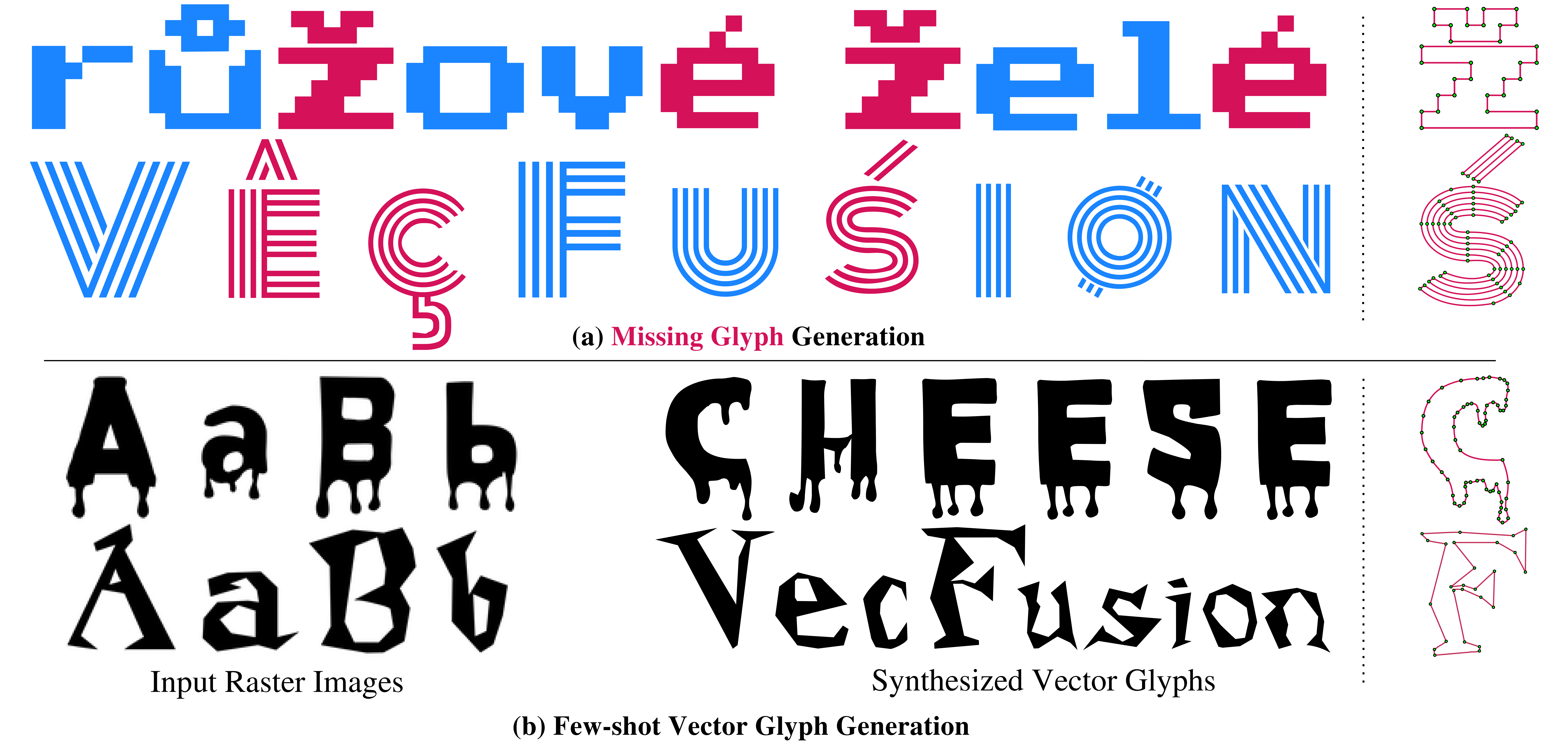}
 \vspace{-2mm}
 \captionof{figure}
  {
    \rev{We present \methodname, a generative model for vector fonts. (a) \methodname generates  missing glyphs in incomplete fonts. \textcolor{blueteaser}{Blue glyphs} are glyphs that exist in the fonts. \textcolor{redteaser}{Red glyphs} are missing glyphs generated by our method. On the right, we show generated control points as circles on selected glyphs. (b) \methodname generates vector glyphs given a few exemplar (raster) images of glyphs. Our method generates precise, editable vector fonts whose  geometry and control points are learned to match the target font style.}
    \label{fig:teaser}
   }   
   \vspace{3mm}
}]
%\maketitle

%%%%%%%%% ABSTRACT
\begin{abstract}
\vspace{-5mm}
We present \methodname, a new neural architecture that can generate vector fonts with varying topological structures and precise control point positions. 
Our approach is a cascaded diffusion model which consists of a raster diffusion model followed by a vector diffusion model. 
The raster model generates low-resolution, rasterized fonts with auxiliary control point information, capturing the global style and shape of the font,
while the vector model synthesizes vector fonts conditioned on the low-resolution raster fonts from the first stage.
To synthesize long and complex curves, our vector diffusion model uses a transformer architecture and a novel vector representation that enables the modeling of diverse vector geometry and the precise prediction of control points. 
Our experiments show that, in contrast to previous generative models for vector graphics, our new cascaded vector diffusion model generates higher quality vector fonts, with complex structures and diverse styles.

\end{abstract}

%\blfootnote{$^*$Equal contribution}

%%%%%%%%% BODY TEXT
% \vspace{4mm}
\section{Introduction}
\label{sec:intro}
\blfootnote{
~$^*$Equal contribution
\\
~Project website: \textcolor{red}{https://vikastmz.github.io/VecFusion/}}
Vector fonts are extensively used in graphic design, arts, publishing, and motion graphics. 
As opposed to rasterized fonts, vector fonts can be rendered at any resolution without quality degradation, and can be edited via intuitive control point manipulation. 
However, authoring high-quality vector fonts remains a challenging and labor-intensive task, even for expert designers.
Recent approaches \cite{lopes2019learned, carlier2020deepsvg, wang2023deepvecfont} use VAEs or autoregressive models to automatically synthesize vector fonts, but they often struggle to capture a diverse range of topological structures and glyph variations, due to the inherent ambiguity of vector curves.
As a result, they frequently create artifacts and imprecise control point positions, compromising the overall quality and editability of the synthesized fonts.

In this work, we leverage recent advances in \emph{raster} generative models, to design a generative model for \emph{vector} fonts.
Such a generative model has a number of real world applications, such as glyph completion, few-shot style transfer, and font style interpolation. 
However, training vector domain generative models is not straightforward: the irregular data structure of vector graphics prevents naive applications of commonly used CNN-based architectures. 
Furthermore, there is an inherent ambiguity in vector representations: infinitely many control points configurations can produce the same glyph, but not all configurations are equivalent.
In particular, designers carefully create control points so that the font can be intuitively edited. 
Generated vector fonts should follow a similar design goal.

To address the above-mentioned challenges, we propose a novel two-stage diffusion model, called \methodname, to generate high-quality vector fonts.
Our pipeline is a cascade of a raster diffusion model followed by a vector diffusion model.
The raster diffusion model gradually transforms a 2D Gaussian noise map to a target raster image of the glyph at low resolution, conditioned on a target glyph identifier and a target font style.
It also generates auxiliary raster fields to drive the placement of vector control points in the next stage.
The second stage is a vector diffusion model conditioned on the raster outputs from the first stage.
It is trained to ``denoise'' a noisy vector glyph representation into structured curves representing the glyph. 

\paragraph{Contributions.}
This papers makes several contributions.
First, we present a novel two-stage cascaded diffusion model for high-quality vector fonts generation.
This cascading process allows us to effectively ``upsample'' low-resolution raster outputs into a vector representation.
We introduce a new mixed discrete-continuous representation for control points, which allows the vector diffusion model to automatically predict the number of control points and paths to use for a glyph, as well as their position. 
We show that diffusion models can effectively ``denoise'' in this new representation space.
Moreover, to capture long-range dependencies and accommodate the irregular nature of vector representations, we introduce a transformer-based vector diffusion model.
\rev{Finally, we show that \methodname synthesizes fonts with much higher fidelity than state-of-the-art methods evaluated on datasets with diverse font styles.}

\section{Related Work}
\label{sec:relwork}

\paragraph{Generative vector graphics.}
Significant work has been invested in generative modeling of vector graphics, using VAEs~\cite{kingma2013auto,lopes2019learned}, sequence-to-sequence models like RNNs~\cite{ha2017neural} or transformers~\cite{ribeiro2020sketchformer}.
Recent approaches employ hierachical generative models~\cite{carlier2020deepsvg}, while others bypass the need for direct vector supervision~\cite{reddy2021im2vec}, using a differentiable rasterizer~\cite{li2020differentiable}.

\vspace{-1mm}

\vspace{-1mm}

\paragraph{Font generation.}
Due to their ubiquity and central role in design, fonts have received special attention and dedicated synthesis methods. 
\rev{Many methods learn to generate \emph{raster} fonts from a large set of reference glyphs \cite{jiang2019scfont, gao2020gan} or a few exemplar images ~\cite{cha2020few, park2021multiple, tang2022few, kong2022look, azadi2018multi, gao2019artistic}. These methods produce visually appealing raster fonts in a variety of styles, but cannot generate vector outputs, thus they are limited by resolution and pixelization artifacts.
In the task of \emph{vector} font generation, early approaches use morphable template models~\cite{suveeranont2010example}, or manifold learning to enable interpolation/extrapolation of existing fonts~\cite{campbell2014learning}, while recent methods use deep generative models~\cite{lian2018easyfont,wang2021deepvecfont}.}
The first generation of deep learning solution sometimes generated glyphs with strong distortion and visual artifacts. 
Methods like DeepVecFont-v2~\cite{wang2023deepvecfont} improve the synthesis quality using a transformer architecture.
Although these methods can generate visually pleasing vector fonts, effectively modeling a diverse distribution of glyphs and topologies remains a challenge. 
DeepVecFont-v2, for instance, only supports a limited number of glyphs (52 characters). 

\paragraph{Diffusion models.}
To address the challenges in vector field design, we leverage diffusion models \cite{ho2020denoising} for their ability to model diverse and complex data distributions. 
\rev{Unlike previous methods \cite{das2023chirodiff, wangsketchknitter} that use CNN or RNN-based vector diffusion models, our approach uses a transformer-based vector diffusion model to handle long-range dependencies inherent to complex vector glyphs.}
Furthermore, our two-stage raster--vector approach and novel vector representation enable precise B\'ezier curve prediction on challenging artist-designed font datasets. 

\vspace{-2mm}
\paragraph{Cascaded diffusion.}
Cascaded diffusion models \cite{ho2022cascaded} have achieved impressive synthesis quality across various domains, including images~\cite{saharia2022photorealistic,balaji2022ediffi},
videos~\cite{ho2022imagen,blattmann2023align} and 3D~\cite{lin2022magic3d,hui2022neural}.
In the same spirit, we introduce a cascaded diffusion model for high-quality vector font generation.

\paragraph{Image vectorization.}
\rev{Image vectorization approaches \cite{tian2022survey, polyvec, ma2022towards} output a vector graphics representation from a raster image. Dedicated for line drawing vectorization, many learning-based methods \cite{kim2018semantic, puhachov2021keypoint, mo2021general} have been proposed. 
Although these methods can produce high-quality vector graphics, they often create redundant or imprecise control points and fail to produce high-fidelity results on low-resolution raster images. 
Our diffusion model can generate precise vector geometry from low-resolution raster images, also providing a new perspective for image vectorization.}

\section{Method}

\begin{figure}[t!]
\centering
\includegraphics[width=0.97\linewidth]{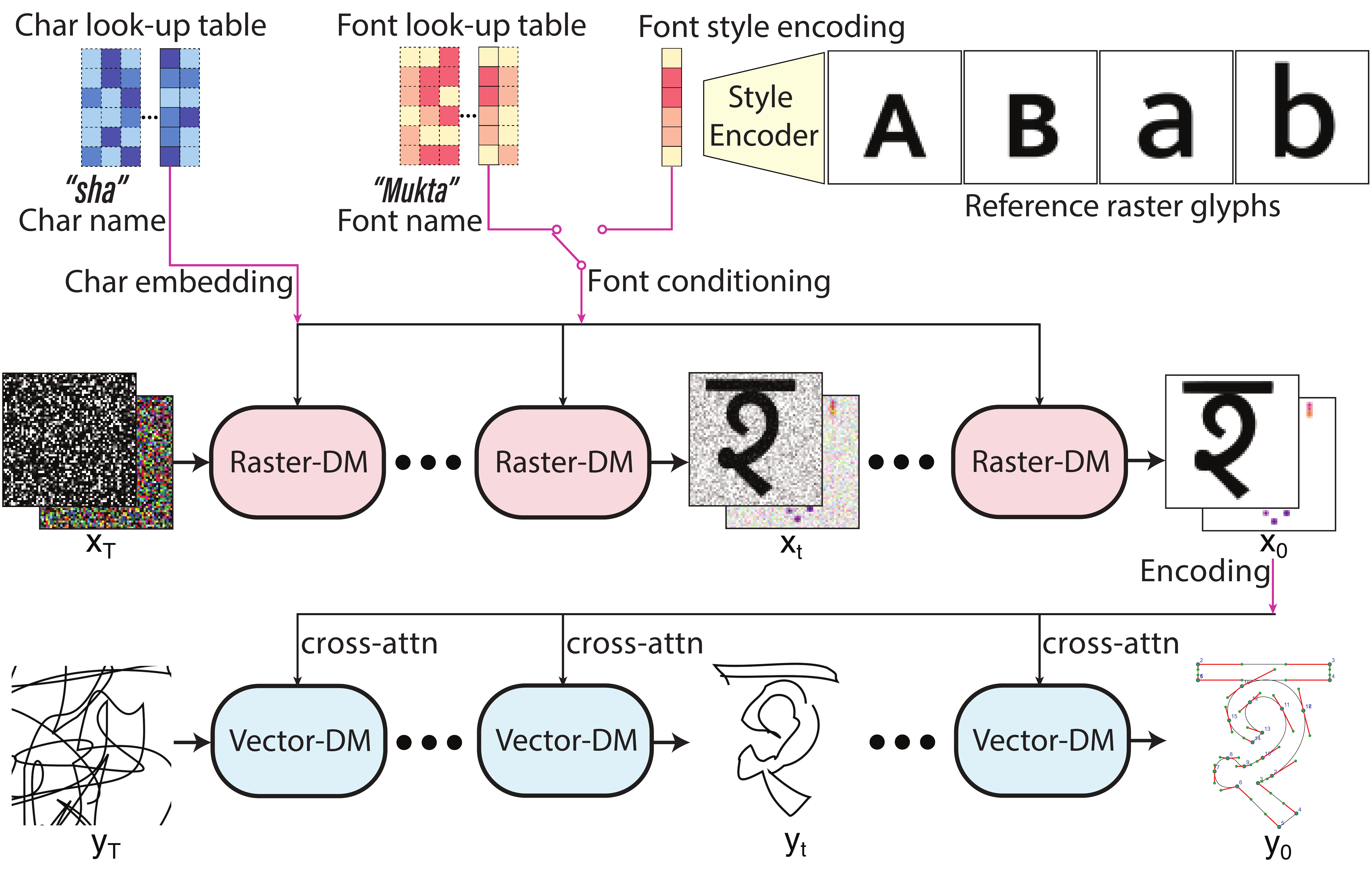}
\vspace{-4mm}
\caption{\rev{Overview of the \methodname's cascade diffusion pipeline. Given a target character and font conditioning, our raster diffusion stage (``Raster-DM'')  produces a raster image representation of the target glyph in a series of denoising steps starting with a noise image. The raster image is encoded and input to our vector diffusion stage (``Vector-DM'') via  cross-attention. The vector diffusion stage produces the final vector representation of the glyph also in a series of denoising steps starting with a noise curve representation.}
}
\label{fig:overview}
\vspace{-1.5em}
\end{figure}

\paragraph{Overview.} The goal of \methodname is to automatically generate vector graphics representations of glyphs. The input to our model is the Unicode identifier for a target character, also known as \emph{code point}, and a target font style. The target font style can be specified in the form of a few representative raster images of other glyphs in that style or simply by the font style name. Figure \ref{fig:overview} shows an example of the generated vector representation for the glyph corresponding to the input target letter ``sha'' of the Devanagari alphabet and the target font style ``Mukta''. 
Our method is trained once on a large dataset of glyphs from various font styles.
Once trained, it can generate glyphs not observed during training.
Our model has several applications: generate missing glyphs in incomplete fonts, synthesize novel fonts by transferring the style of a few exemplar images of glyphs, or interpolate font styles. 

\paragraph{Output vector representation.} The generated vector representation for a glyph is in the form of ordered sequences of control points in cubic B\'ezier curve paths commonly used in vector graphics. 
Control points can be repeated in the generated sequences to manipulate the continuity of the vector path. Our method learns to generate an appropriate number of vector paths, control points, and point repetitions tailored to each character and font style. In addition, it learns the proper ordering of control points for each path, including where first and last control points are placed, since their placement patterns often reflect artist's preferences.

\paragraph{Pipeline.} Our method consists of a two-stage cascade. 
\rev{In the first stage (raster diffusion model, or in short ``Raster-DM'', Figure \ref{fig:overview}), conditioned on the target character code point and font style, our method initiates a reverse diffusion process to generate a raster image.} The generated raster image captures the shape and style of the target glyph at low resolution. Additionally, we generate an auxiliary set of \emph{control point fields} encoding information for control point location, multiplicity, and ordering. In the second stage (vector diffusion model or ``Vector-DM'', Figure \ref{fig:overview}), our method proceeds by synthesizing the vector format capturing fine-grained placement of control points guided by \rev{the raster glyph and control point fields generated in the first stage.} We observed that this two-stage approach results in generating higher-fidelity fonts compared to using diffusion in vector space directly or without any guidance from our control point fields. In the next sections, we discuss our raster and vector diffusion stages in more detail.

\subsection{Raster diffusion stage}

Given a target character identifier and font style, the raster diffusion stage creates a raster image $\bx_0$ encoding information about the target glyph in pixel space (Figure \ref{fig:overview}, ``Raster-DM'').  
This is performed through a diffusion model that gradually transforms an image $\bx_T$ sampled from a unit Gaussian noise distribution towards the target raster image $\bx_0$ in a series of $T$ denoising steps.  At each  step $t=1...T$, a trained neural network executes the transition $\bx_{t} \rightarrow \bx_{t-1}$ by predicting the noise content to be \rev{removed} from the image $\bx_{t}$. \rev{This denoiser network is conditioned on the input character identifier and font style. In the next paragraphs, we explain the encoding of the input character codepoint and font style}, the target raster image, the denoiser network, and finally the training and inference processes of this stage. 

\paragraph{Character identifier embeddings.} Inspired by similar approaches in NLP to represent words \cite{vaswani2017attention}, we create a one-hot vector representation for all unique character codepoints available in our dataset. Given a target character's codepoint, its one-hot vector representation is mapped to a continuous embedding $\bg$ through a learned look-up table. The look-up table stores embeddings for all codepoints available in our dataset and retrieves them using the one-hot vector as indices. 

\paragraph{Font style conditioning.} \rev{To encode the font style, we experimented with two approaches depending on the application. To generate missing glyphs in incomplete fonts, we create a one-hot vector representation for all font styles available in our dataset. Given a target font style, its one-hot vector is mapped to a continuous embedding $\bff$ through a learned look-up table as above. 
To generate glyphs conditioned on a few exemplar images, we concatenate the input images channel-wise and pass them through a convnet to get a font style feature map $\bff$ (see supplement for more details.) } 

\setlength{\columnsep}{10pt}
\begin{wrapfigure}{R}{0.22\columnwidth}
 \vspace{-2mm} 
  \includegraphics[width=1\linewidth]{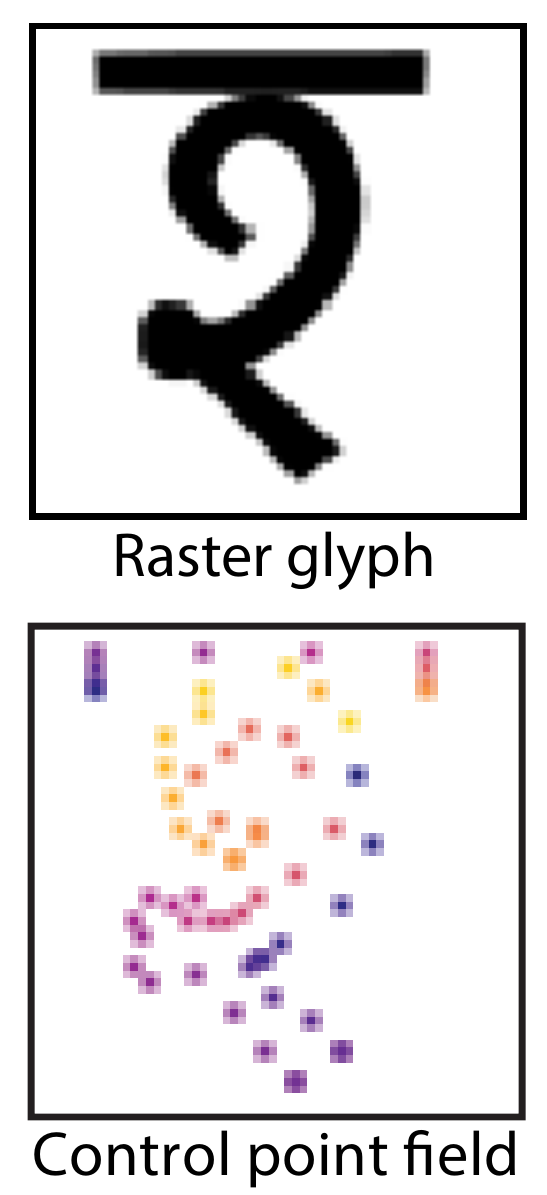}
  \vspace{-8mm} 
  \caption{Target $\bx_0$}
  \vspace{-4mm}
\label{fig:target_image}
\end{wrapfigure}
\paragraph{Target raster image.} The target $\bx_0$ produced in the raster diffusion stage is a $N \times N$ image made of the following  channels:

\noindent (a) the first channel is composed of an image representing a grayscale rasterized image of the target glyph (Figure \ref{fig:target_image}, top).

\noindent (b) the rest of the channels store \emph{control point fields} (Figure \ref{fig:target_image}, bottom), whose goal is to encode information about the control point location, multiplicity, and ordering. During training, this control point field is created by rendering each control point as a Gaussian blob centered at the 2D coordinates  ($x$,$y$) of the control point. The coordinates are normalized in $[0,1]^2$. We also modulate the color of the blob based on (a) the index of the control point in the sequence of control points of its vector path (e.g., first, second, third etc control point), and (b) its multiplicity. A look-up function is used to translate the ordering indices and multiplicities of control points to color intensities. In our implementation, we use $3$ channels for this control point field, which can practically be visualized as an RGB image (Figure \ref{fig:target_image}, bottom). These channels are concatenated with the raster image of the glyph, forming a $4$-channel image.

\paragraph{Raster denoiser.} The denoiser is formulated as a UNet architecture \cite{dhariwal2021diffusion}. The network takes the $4$-channel image $\bx_t$ as input and is conditioned on the embedding of time step $t$. 
\rev{Following \cite{rombach2022high}, we add the character's codepoint embedding $\bg$ to time step embedding and pass it to each residual block in the UNet. 
For the font style conditioning, we add it to the time step embedding if it is a single embedding. If the font style is encoded as a spatial feature map, following \cite{rombach2022high}, we flatten the feature map and inject it to the UNet via cross-attention.
}

The denoiser network predicts the per-channel noise component of the input image, which is also a $4$-channel image (see supplement for more details.).

\paragraph{Training loss} The network is trained to approximate an optimal denoiser under the condition that the images $\bx_1, \bx_2, ... \bx_{T}$ are created by progressively adding Gaussian noise to the image of the previous step \cite{ho2020denoising}: 
\mbox{$q(\bx_t | \bx_{t-1}) = \mathcal{N}\big( \bx_t; \sqrt{(1-\beta_t)} \bx_{t-1}, \beta_t \bI \big)$}, where $\beta_t$ represents the variance of the Gaussian noise added at each step. The image $\bx_T$ converges to a unit Gaussian distribution as $T \rightarrow \infty$, or practically a large number of steps \cite{ho2020denoising}. 
 Following \cite{ho2020denoising}, we train the denoiser network with the training objective $||\bepsilon(\bx_t, t, \bff, \bg) - \bepsilon||^2$ i.e. the mean-squared error loss between the added training noise $\bepsilon$ at each step and the predicted noise $\bepsilon(\bx_t, t, \bff, \bg)$ from the network. The loss is used to train the denoiser and the look-up tables.

\paragraph{Inference.} At test time, given sampled unit Gaussian noise $\bx_T$, a target character embedding $\bg$ and font style conditioning $\bff$, the network is successively applied in $T$ steps to generate the target raster image. 

\paragraph{Implementation details.} In all our experiments, we use the following hyperparameters. Following \cite{nichol2021improved}, we set the
number of diffusion steps $T$ to 1000 and used cosine noise schedule in the forward diffusion process. Training takes $5$ days on 8 A100 GPUs.
We used the AdamW optimizer \cite{loshchilov2017decoupled} with learning rate $3.24 \cdot 10^{-5}$. 
The feature embeddings for character identifiers are set to be $896$-dimensional. 
The control points are rendered as Gaussian blobs with radius of $2$ pixels. The raster image resolution is set to $64 \times 64$. Lower resolutions cause increasing overlaps between the rendered blobs, making the control point field more ambiguous. Increasing the resolution increases the computational overhead for the raster denoiser. The above resolution represented a good trade-off, as we practically found in our experiments. 
As mentioned above, we use $3$ channels to encode control point ordering and multiplicity as colors. We practically observed that $3$ channels were enough to guide the vector diffusion stage. 
Depending on the dataset, fewer channels could be used instead e.g., in cases of glyphs with few control points, or no multiplicities. In general, these hyperparameters can be adjusted for different vector graphics tasks -- our main point is that the raster images and fields are useful as guidance to produce high-fidelity vector fonts, as shown in our ablation.

\subsection{Vector diffusion stage}

Given the raster image generated in the previous stage, the vector diffusion stage creates a tensor $\by_0$ representing the target glyph in vector graphics format (Figure \ref{fig:overview} ``Vector-DM''). 
The reverse diffusion process gradually transforms a noise tensor $\by_T$ sampled from a unit Gaussian noise distribution towards a tensor $\by_0$ in a series of denoising steps.
In this domain, the noise represents noise on the \emph{spatial position} and \emph{path membership} of the control points, rather than the intensity of the pixel values as in the raster domain.
In the next paragraphs, we explain the tensor representation, the denoiser,  training and inference of this stage.

\begin{figure}[t!]
\centering
\includegraphics[width=0.97\linewidth]{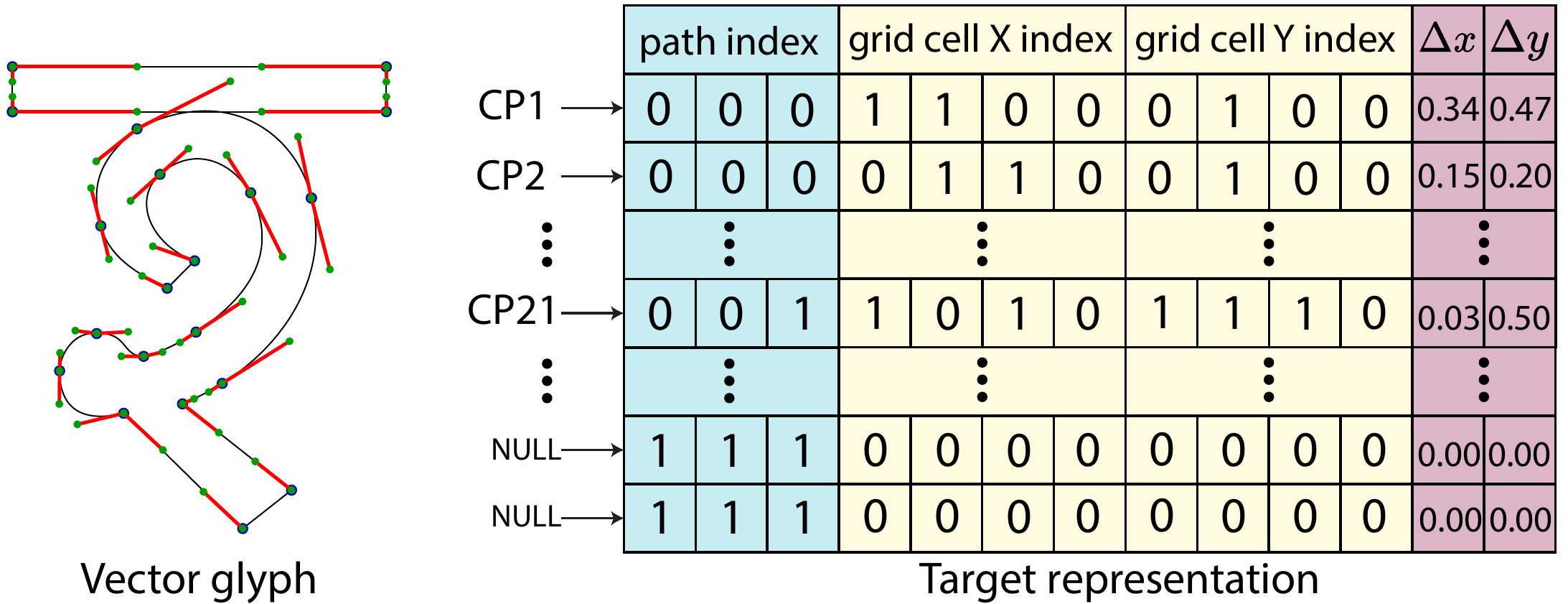}
\vspace{-2mm}
\caption{\emph{Target tensor representation $\by_0$.} Our vector diffusion model ``denoises'' this tensor representation which includes both path membership and spatial position for control points. The discrete values (path membership, grid cell coordinates) are denoised in the continuous domain and then discretized. The control point locations are computed from the predicted grid cell coordinates plus continuous displacements $(\Delta x, \Delta y)$ from them.}
\vspace{-3mm}

\label{fig:representation}
\end{figure}

\paragraph{Target tensor.}
The target tensor $\by_0$ is a $M \times D$ tensor (Figure \ref{fig:representation}), where $M$ represents an upper bound to the total number of control points a glyph can have. Each entry in the tensor contains a $D$-dimensional representation of a control point. Specifically, each entry stores the following information:

\noindent (a) the index of the vector path the control point belongs to i.e, its path membership. During training, each vector path is assigned a unique index. Since the vector paths can be re-ordered arbitrarily without changing the resulting glyph, to reduce unnecessary variability during learning, we lexigraphically sort vector paths using the coordinates of their control point closest to the top-left corner of the glyph raster image as sorting keys. Following \cite{chen2023analog}, the resulting sorted path index is converted to binary bits. For each control point entry, we store the binary bits of its vector path. A null entry (i.e., all-one bits) is reserved for entries that do not yield control points -- in this manner, we model vector fonts with a varying number of control points and paths.

\noindent (b) the index of the grid cell containing the control point. We define a coarse $P$ $\times$\ $P$ grid over the image, with $P^2$ corresponding grid cell centroids. We assign each control point to the grid cell that has the closest centroid. \rev{In case the control point lies on the boundary of two cells, we use a round operation that assigns it in the second cell.} Similar to path membership, the grid cell index is converted to binary bits. For each control point entry, we store the binary bits of its assigned grid cell.

\noindent (c) the continuous coordinates of the control point expressed relative to the center of the grid cell 
it belongs to. These are two continuous values capturing the  location of each control point. We found that capturing control point locations relative to cell centers achieves the best performance. Since the generated raster control point field (approximately) highlights regions storing control points, mapping the control point field to discrete cell indices plus small continuous residuals, or displacements, 
is an easier task and reduces the continuous coordinate variability needed to be captured by the model. 

\vspace{-1em}
\paragraph{Denoiser.} The denoiser for this stage is formulated as an encoder-only transformer \cite{devlin2018bert}, which takes the tensor $\by_t$ as input and is conditioned on the embedding of time step $t$, and the generated raster image $\bx_0$ from the raster diffusion model. 
We use a ConvNet to encode the raster image $\bx_0$ into high-dimensional features, which are input to the transformer via cross-attention similar to \cite{rombach2022high}.
The transformer predicts the noise content as a $M$ $\times$ $D$ tensor at each step.

\paragraph{Training loss.} We train the denoiser network according to mean-squared error loss between training noise and predicted one at sampled time steps: $||\bepsilon(\by_t, \bx_0, t) - \bepsilon||^2$. 

\paragraph{Inference.} At test time, given a sampled tensor $\by_T$ from unit Gaussian noise and a generated raster image $\bx_0$ of the previous stage, the denoiser network is applied in a series of $T$ steps to generate  the target tensor $\by_0$. Following the Analog Bits approach \cite{chen2023analog}, the discrete binary bits in the target tensor representation are modeled as real numbers. These are simply thresholded to obtain the final binary bits at inference time. 
Given the predicted path membership, we create a set of vector paths according to the largest generated control path index number. Each non-null entry in the generated tensor yields a control point. The control points are implicitly ordered based on their entry index. The location of the control point is defined as the coordinate center of the assigned cell in the generated tensor plus the predicted relative displacement. Given this generated information, we directly reconstruct the vector paths without requiring any further refinement or post-processing. 

\paragraph{Implementation details.} 
In our implementation, we set the upper bound for the number of control points to $M=256$, which was sufficient for the datasets we experimented with. We use $3$ bits to represent the path membership, which can support up to $7$ distinct vector paths. This was also sufficient for the datasets in our experiments. 
We set $P$ to $16$, resulting in $256$ grid cells which can be represented by $8$ binary bits. Together with the two-dimensional relative displacement, the final dimension of our target tensor $D$ is $13$ in our experiments.
Similar to our raster diffusion model, we set the number of diffusion steps $T$ to 1000, used cosine noise schedule, and the AdamW optimizer \cite{loshchilov2017decoupled} with learning rate $3.24 \cdot 10^{-5}$. Training is done separately from the raster diffusion stage and takes $5$ days on 8 A100 GPUs. During testing, we use the DDPM sampler \cite{ho2020denoising} with $1000$ steps. Generating a glyph by executing both stages takes around 10 seconds on a A100 GPU.

\section{Results}
\label{sec:results}

In this section, we present experiments in three different application scenarios for our method. In the first scenario, we address the problem of \emph{missing glyph generation} for a font (Section \ref{sec:unicode-completion}). Many users often experience a frustrating situation when they select a font they prefer, only to discover that it lacks certain characters or symbols they wish to utilize. This issue is particularly prevalent when it comes to non-Latin characters and mathematical symbols. As a second application, we apply our method for \emph{few-shot font style transfer} (Section \ref{sec:font-style-transfer}), where the desired font is specified in the form of a few exemplar raster glyphs, and the goal is to generate vector glyphs in the same font. Finally, we discuss \emph{interpolation of font styles} (Section \ref{sec:interpolation}) i.e., generate glyphs whose style lies in-between two given fonts.

\subsection{Generation of missing Unicode glyphs}
\label{sec:unicode-completion}
\label{sec:missing-glyph-generation}

Existing public datasets or benchmarks to evaluate glyph generation are limited to a specific alphabet (e.g., Latin). Below we discuss a new dataset for evaluating generation of glyphs across different languages, math symbols, and other signs common in the Unicode standard. Then 
we discuss comparisons, ablation study, metrics for evaluation, and results.

\paragraph{Dataset.} We collected a new dataset of $1424$ fonts from Google Fonts.  
The dataset contains $324K$ glyphs, covering $577$ distinct Unicode glyphs in various languages (e.g., Greek, Cyrillic, Devanagari), math symbols and other signs (e.g. arrows, brackets, currency).
We randomly partition the dataset into $314K$-$5K$-$5K$ glyphs for training, validation, and testing respectively. 

\paragraph{Comparison.}
We compare with ``ChiroDiff'' \cite{das2023chirodiff}, which 
applies diffusion models to generate Kanji characters as polylines. Their method uses a set-transformer \cite{lee2019set} to obtain a latent embedding from a 2D point set as the input condition. We replace their input condition to their diffusion model with the embeddings of characters and fonts using look-up tables, as done in our raster diffusion model. 
We trained and tuned their method, including the embeddings, to predict B\'ezier  curve control points using our dataset, as in our method. 

\paragraph{Ablation.}
In addition, we evaluate the following alternative variants of our method: (a) \textbf{\textit{Vector only:}} in this ablation, we remove the raster diffusion model and use the vector diffusion model only -- in this case, the font and character embeddings are used as input conditions to the vector diffusion model. (b) \textbf{\textit{No control point fields:}} we remove the RGB control point field  from the target raster image of our raster diffusion model -- in this case, we condition the vector diffusion model only on the single-channel raster image of the glyph. (c) \textbf{\textit{Predict continuous coordinates only:}} in the vector diffusion model, instead of predicting 
discrete grid cell indices plus displacements relative to cell centers, we directly predict the absolute coordinates x and y per control point.

\begin{figure*}
    \centering
    \includegraphics[width=0.95\linewidth]{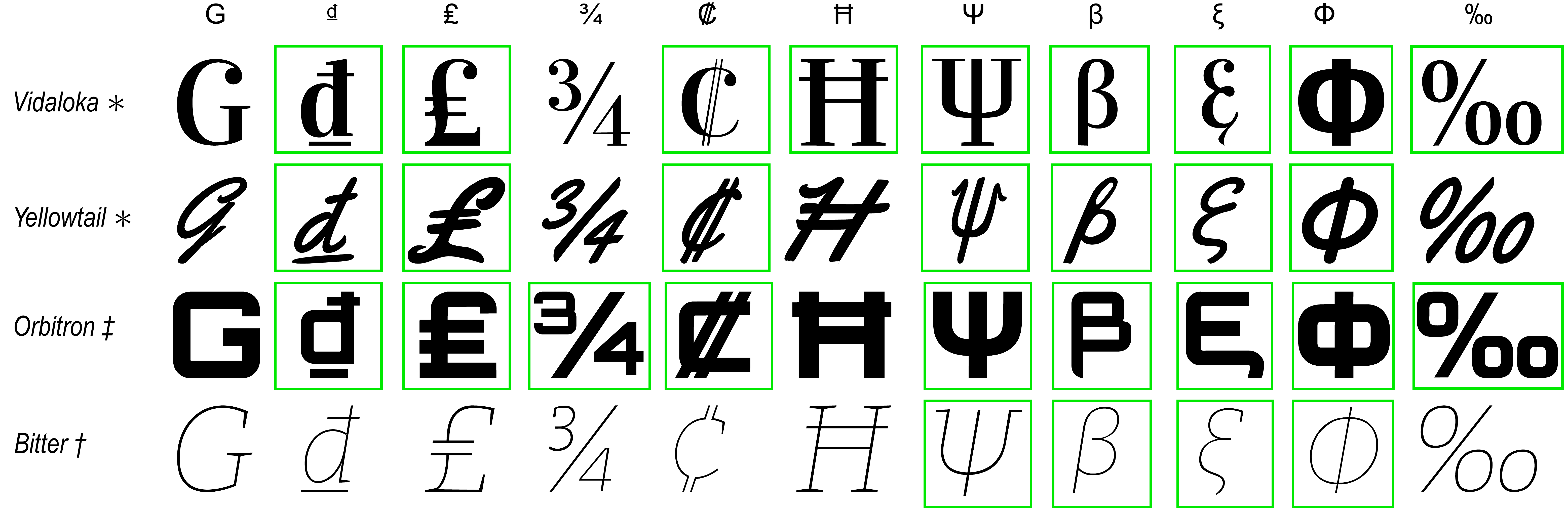}
        \vspace{-3mm}
    \caption{An incomplete font matrix from the Google Font dataset, each row represents a font and all glyphs in one column have the same Unicode. Glyphs in the green boxes are \textcolor{fontgreen}{missing glyphs} generated by our method. $^\ast$: Regular, $^\ddagger$: ExtraBold, $^\dagger$: Italic\-VariableFontWidth.
    \label{fig:more_results} }
    %\vspace{-1mm}
\end{figure*}

\paragraph{Evaluation metrics.}
We compare the generated glyphs with ones designed by artists in the test split. We use the following metrics:

\noindent (a) $L_1$: we compare the image-space absolute pixel differences of glyphs when rasterized. We use the same rasterizer for all competing methods and variants. This reconstruction error was also proposed in \cite{wang2021deepvecfont} for glyph evaluation.
 
\noindent (b) \emph{CD:} we measure the bidirectional Chamfer distance between artist-specified control points and generated ones.

\noindent (c) \emph{ \#cp diff:} we measure the difference between the  number of artist-made control points and predicted ones averaged over all paths.

\noindent  (d) 
\emph{ \#vp diff:} we measure  the difference between the number of artist-specified vector paths and predicted ones.

For all the above measures, we report the averages over our test split.
We propose the last three metrics for comparing glyphs in terms of the control point and path characteristics, which are more relevant in vector font design.

\begin{table}[t]
    \centering
    % \small
    \setlength{\tabcolsep}{2pt}
    \begin{tabular}{l ccccc}
         \toprule
         \textbf{Method} & $L_1$ $\downarrow$ & CD $\downarrow$ & \#cp diff $\downarrow$ & \#vp diff $\downarrow$
         \\
         \midrule
        Ours 
        & \textbf{0.014} & \textbf{0.16}  & \textbf{3.05}  & \textbf{0.03}\\
        % Srivatsan et al. \cite{srivatsan2021scalable}
        Ours (cont. coord. only)
        & 0.020  &  0.18 &  3.30  & \textbf{0.03}\\
        Ours (no cp fields)
        & 0.016 & 0.60  &  12.46  & 0.13 \\
        Ours (vector only)
        & 0.016 & 0.68  & 9.36   & 0.11\\
        \hdashline
        ChiroDiff \cite{das2023chirodiff}
        & 0.044 &  1.66 &  56.37  & 0.77\\
        \bottomrule
    \end{tabular}
    \vspace{-3mm}
    \caption{\rev{Missing glyph generation evaluation on the full test set.}}
    % CD is scaled by $10^2$.
    \label{tab:font_reconstruction}
    \vspace{-2mm}
\end{table}
\begin{table}[t]
    \centering
    % \small
    \setlength{\tabcolsep}{2pt}
    \begin{tabular}{l ccccc}
         \toprule
         \textbf{Method} & $L_1$ $\downarrow$ & CD $\downarrow$ & \#cp diff $\downarrow$ & \#vp diff $\downarrow$
         \\
         \midrule
        Ours 
        & \textbf{0.021} & \textbf{0.35}  & \textbf{3.92}  & \textbf{0.04}\\
        % Srivatsan et al. \cite{srivatsan2021scalable}
        Ours (cont. coord. only)
        & 0.028  &  0.40 &  4.11  & \textbf{0.04}\\
        Ours (no cp fields)
        & 0.026 & 0.83  &  12.90  & 0.16 \\
        Ours (vector only)
        & 0.025 & 0.84  & 9.72   & 0.15\\
        \hdashline
        ChiroDiff \cite{das2023chirodiff}
        & 0.072 &  3.63 &  71.97  & 1.02\\
        \bottomrule
    \end{tabular}
    \vspace{-3mm}
    \caption{\rev{Missing glyph generation evaluation on a more challenging subset of our test set where test glyphs are from different glyph families compared to any glyphs in the training set.}}
    % CD is scaled by $10^2$.
    \label{tab:font_reconstruction_2}
    \vspace{-3mm}
\end{table}
\begin{table}[t]
    \centering
    \small
    \setlength{\tabcolsep}{2pt}
    \begin{tabular}{l ccccc}
         \toprule
         \textbf{Method} & $L_1$ $\downarrow$ & CD $\downarrow$ & \#cp diff $\downarrow$ & \#vp diff $\downarrow$
         \\
         \midrule
        Ours 
        & \textbf{0.069} & \textbf{0.46}  & \textbf{15.05} & \textbf{0.033} \\
        DeepVecFont-v2  \cite{wang2023deepvecfont}
        & 0.098 & 1.05  & 34.84 & 0.052 \\
        DualVector \cite{dualvector}
        & 0.197 & 1.37  & 25.69 & 0.428 \\
         \bottomrule
    \end{tabular}
    \vspace{-3mm}
    \caption{\rev{Few-shot font style transfer evaluation. %\cite{dualvector}
    }}
    % CD is scaled by $10^2$.
    \label{tab:font_style_transfer_rebuttal}
    \vspace{-5mm}
\end{table}

\vspace{-0.2em}
\paragraph{Quantitative Results.} Table \ref{tab:font_reconstruction} shows the quantitative results for ChiroDiff and the alternative variants of our method on the full test set. The full version of our method outperforms ChiroDiff and our reduced variants on all evaluation metrics. 
\rev{We note that a glyph in one family variation e.g., italics might have a different variation in the same font, e.g., bold. Although two different family variations of the same glyph often have largely different control point locations and distributions, we create an even more challenging subset of the test set where we removed all test glyphs that had a different family variation in the same font within the training set. The results on this subset are reported in Table \ref{tab:font_reconstruction_2}. Chirodiff and other variants still have much higher error than ours. This indicates that our two-stage approach, and the mixed discrete-continuous representation of vector paths along with the control point fields are all important to achieve high performance.}

\begin{figure}[t]
\centering
\includegraphics[width=0.97\linewidth]{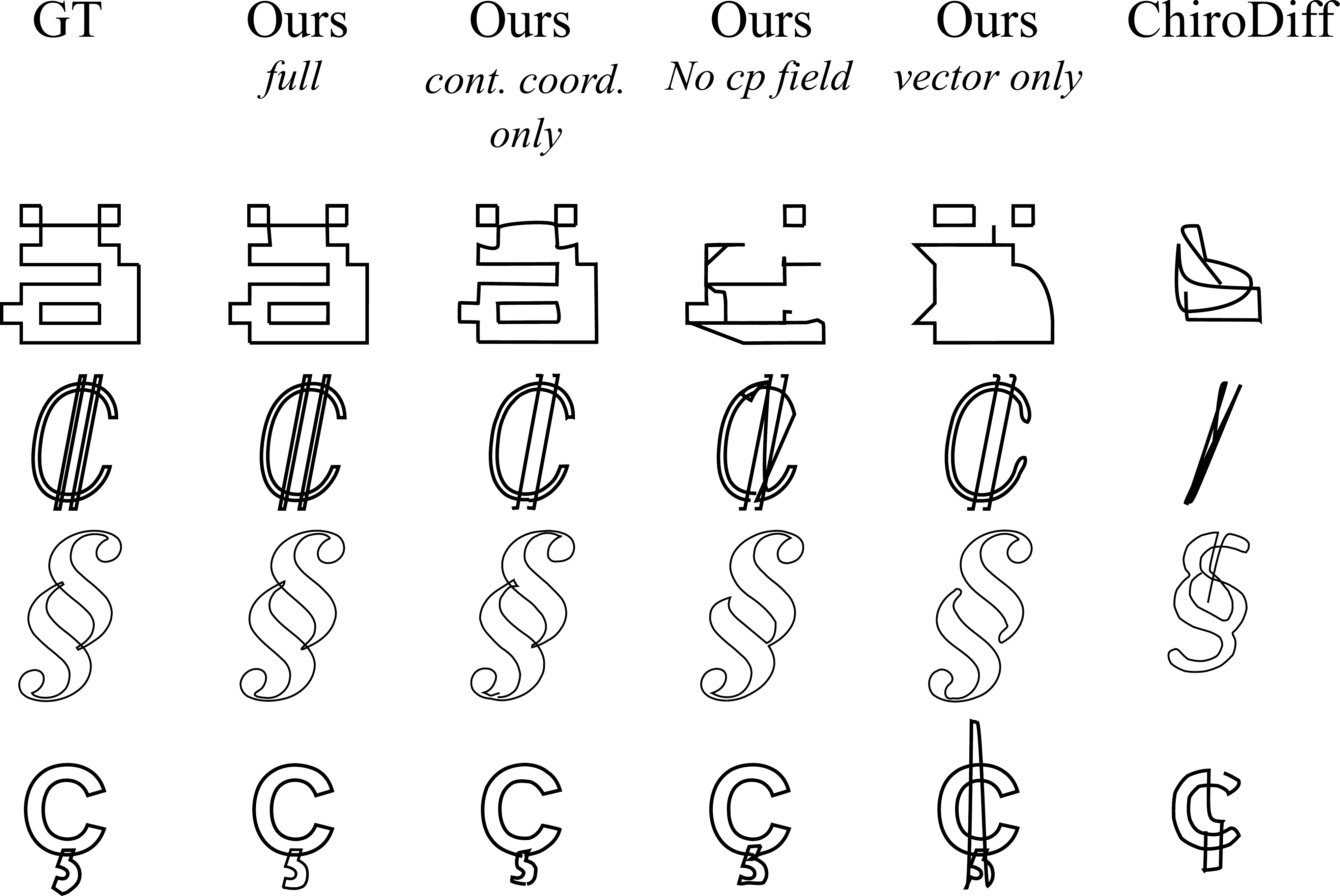}
\vspace{-2mm}
\caption{\rev{Glyph generation results for test cases from the Google font dataset. We compare our method to ChiroDiff \cite{das2023chirodiff} and degraded variants of our method. Our full method is able to generate glyphs that are much closer to artist-made (``ground-truth''/``GT'') ones compared to alternatives. }
}
\label{fig:missing_testset_comp}
\vspace{-3mm}
\end{figure}

\begin{figure}[t]
    \centering
    \setlength{\tabcolsep}{3pt}
    \renewcommand{\arraystretch}{3}

    \includegraphics[width=0.45\textwidth]{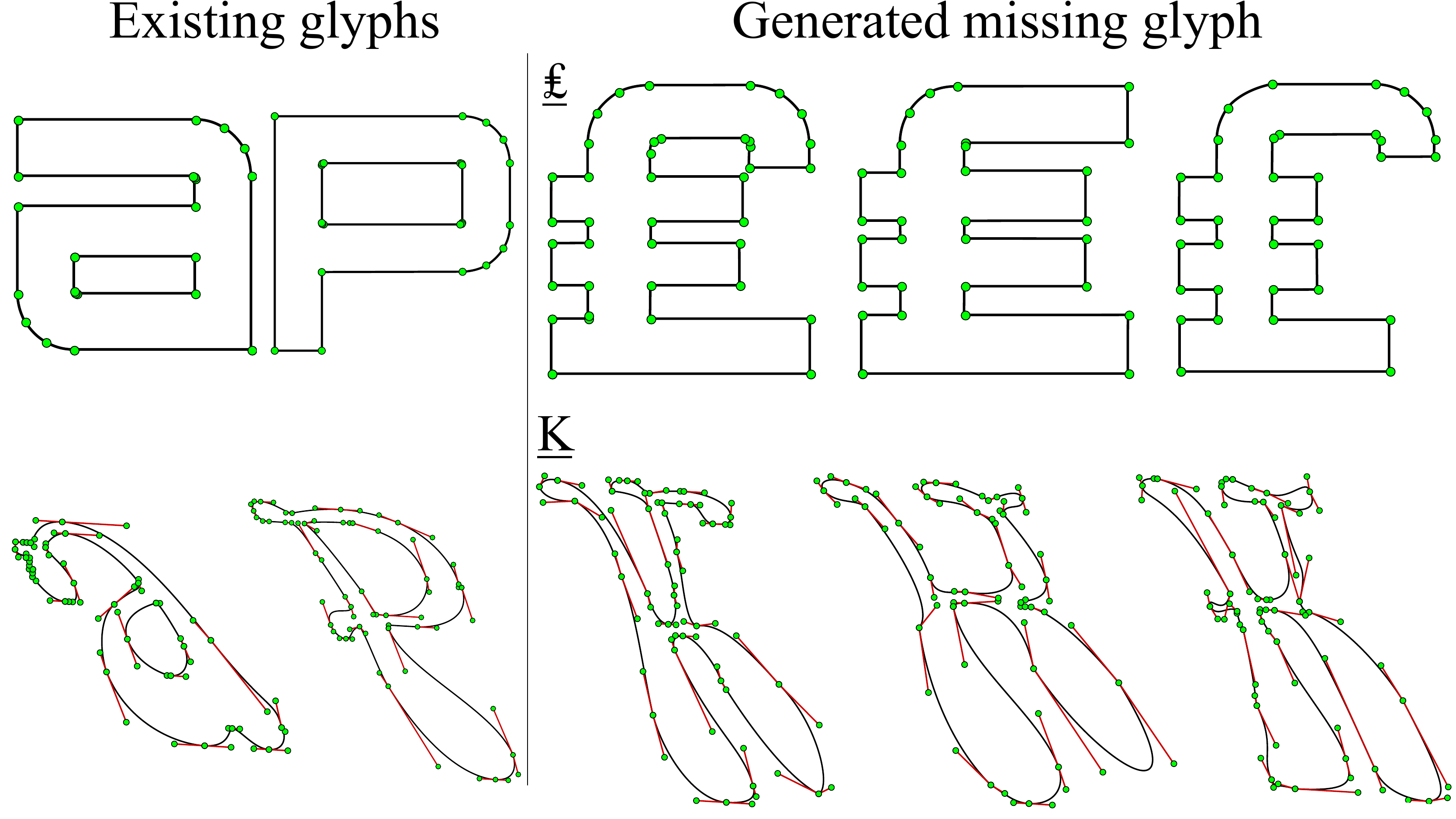}
    \vspace{-3mm}
    \caption{Stochastic sampling: the results are generated with three different random seeds. \label{fig:stochasticity}}
    \vspace{-3mm}
\end{figure}

\begin{figure}[t]
\vspace{-3mm}
    \centering
    \setlength{\tabcolsep}{1pt}
    \renewcommand{\arraystretch}{2}
    \begin{tabular}{llll !{\vrule height 5.1ex}  c c c c}
        \multicolumn{4}{c}{Input reference} &  (a) & (b) & (c) & (d) \vspace{-2mm} \\ 
        % \hline
        % &  & &  \\
\includegraphics[height=0.055\textwidth,width=0.055\textwidth]{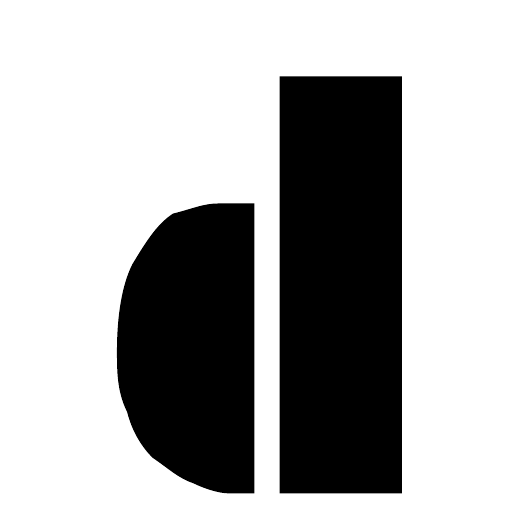} 
&
 \includegraphics[height=0.055\textwidth,width=0.055\textwidth]{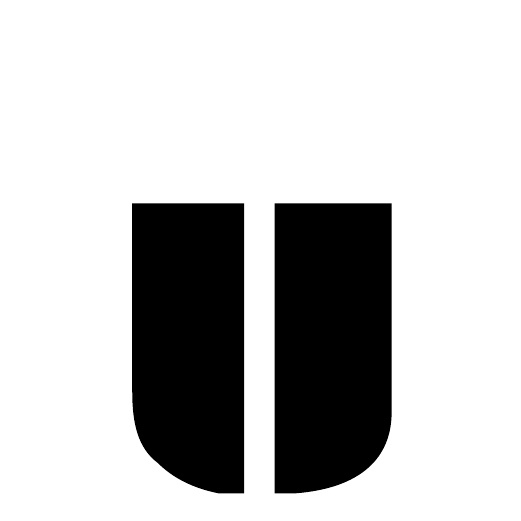} 
&
 \includegraphics[height=0.055\textwidth,width=0.055\textwidth]{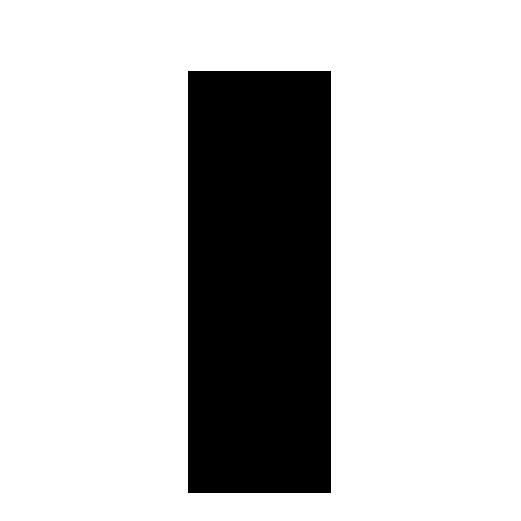} 
&
 \includegraphics[height=0.055\textwidth,width=0.055\textwidth]{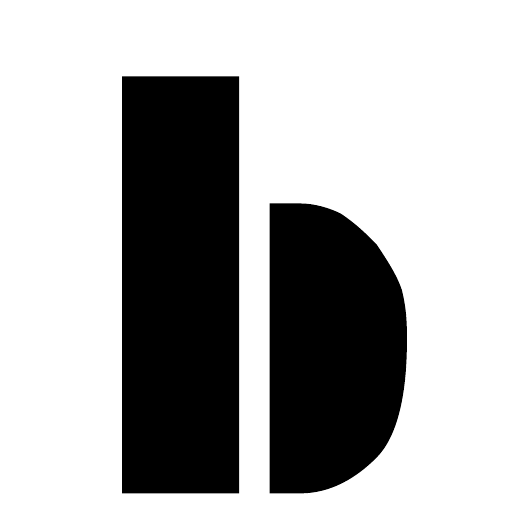} 
&
 \includegraphics[height=0.055\textwidth,width=0.055\textwidth]{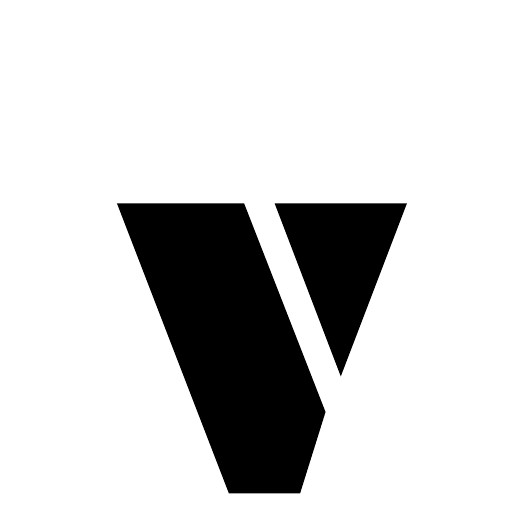} 
&
 \includegraphics[height=0.055\textwidth,width=0.055\textwidth]{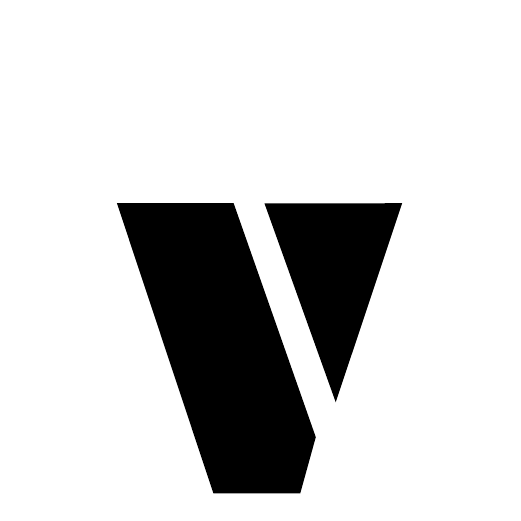} 
&
 \includegraphics[height=0.055\textwidth,width=0.055\textwidth]{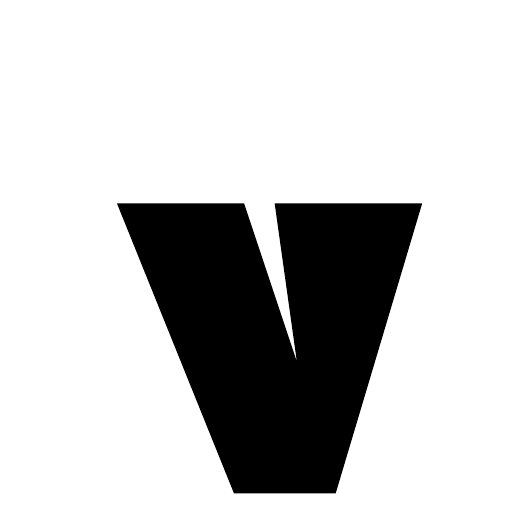} 
 &
 \includegraphics[height=0.055\textwidth,width=0.055\textwidth]{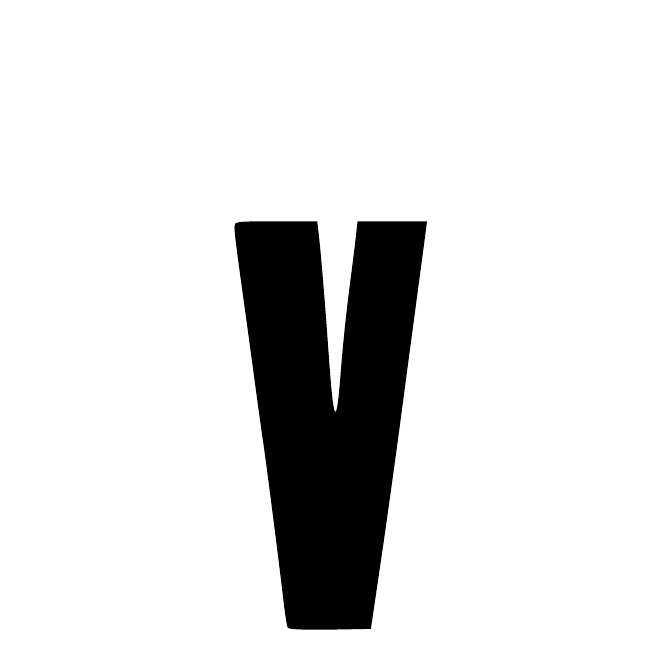} \vspace{-3mm} \\
 
\includegraphics[height=0.055\textwidth,width=0.055\textwidth]{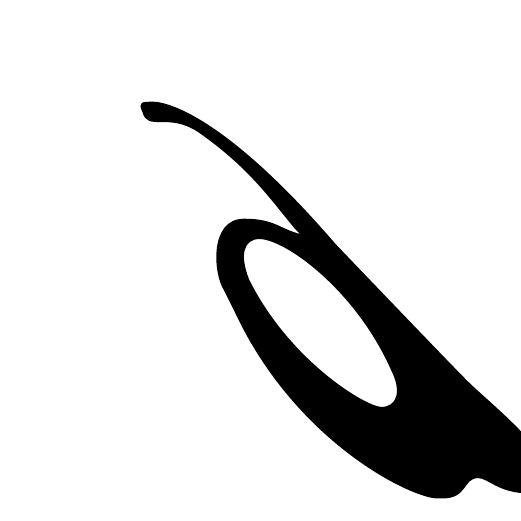} 
&
 \includegraphics[height=0.055\textwidth,width=0.055\textwidth]{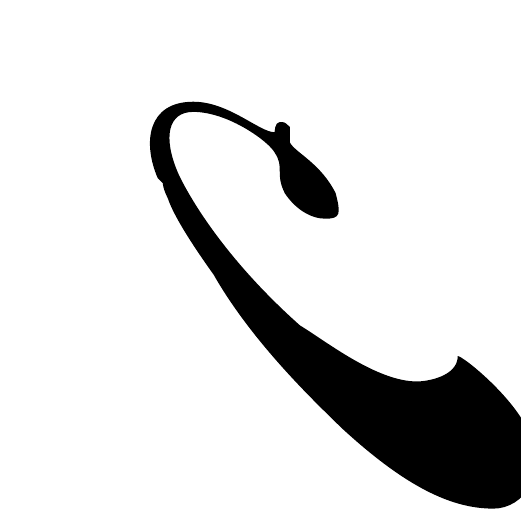} 
&
 \includegraphics[height=0.055\textwidth,width=0.055\textwidth]{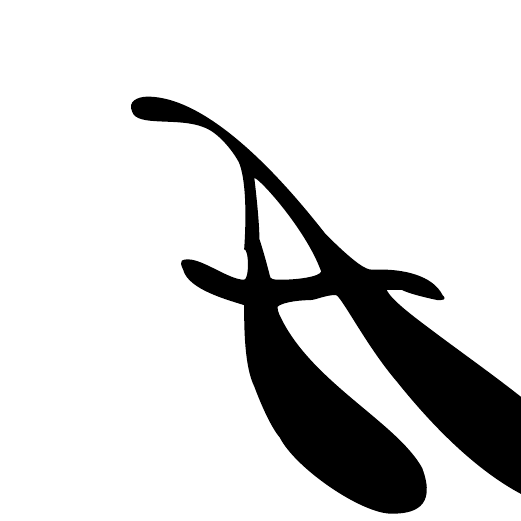} 
&
 \includegraphics[height=0.055\textwidth,width=0.055\textwidth]{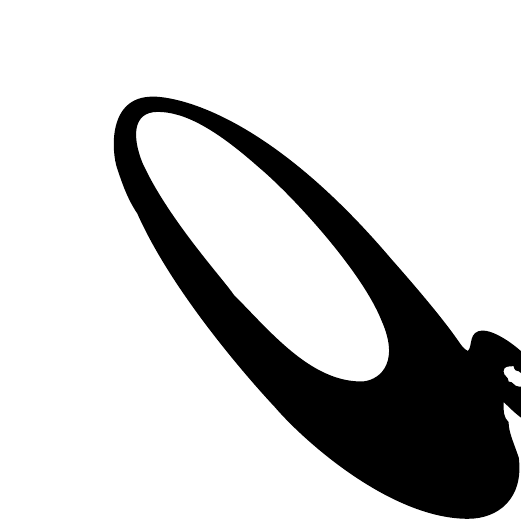} 
&
 \includegraphics[height=0.055\textwidth,width=0.055\textwidth]{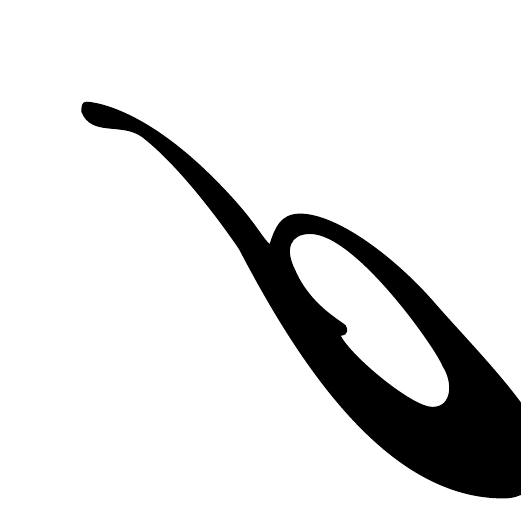} 
&
 \includegraphics[height=0.055\textwidth,width=0.055\textwidth]{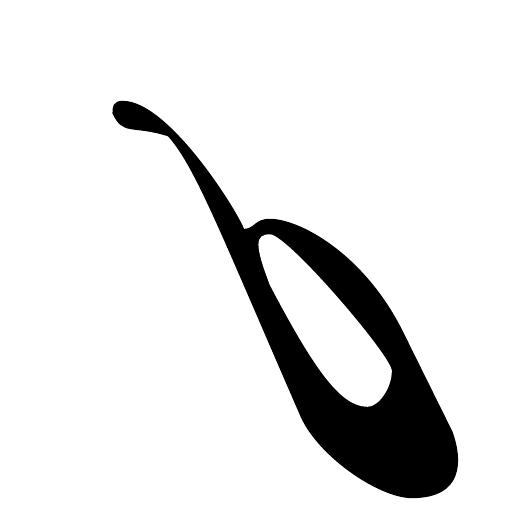} 
&
 \includegraphics[height=0.055\textwidth,width=0.055\textwidth]{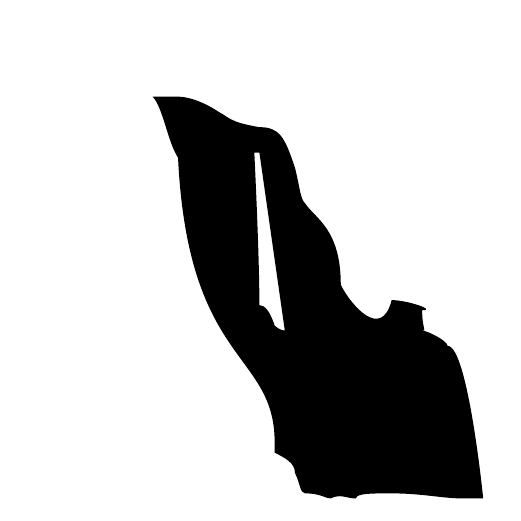}  

 &
 \includegraphics[height=0.06\textwidth,width=0.06\textwidth]{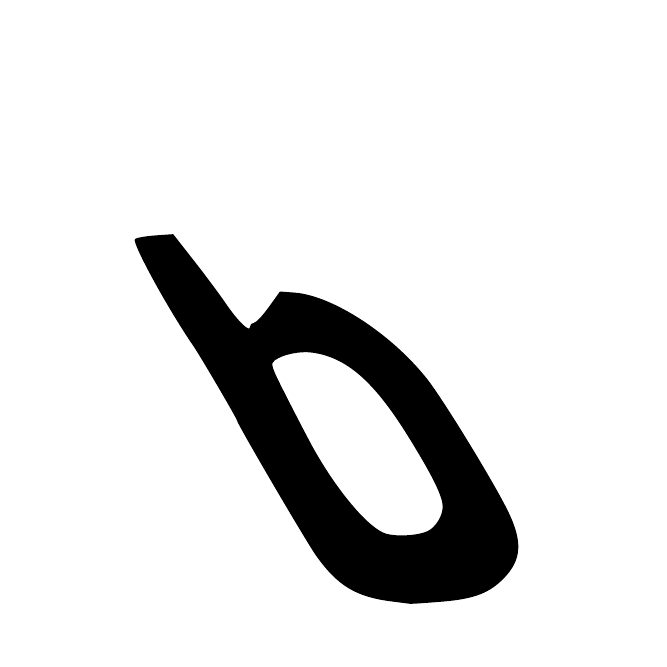} \vspace{-3mm}
 \\
 
 \includegraphics[height=0.05\textwidth,width=0.05\textwidth]{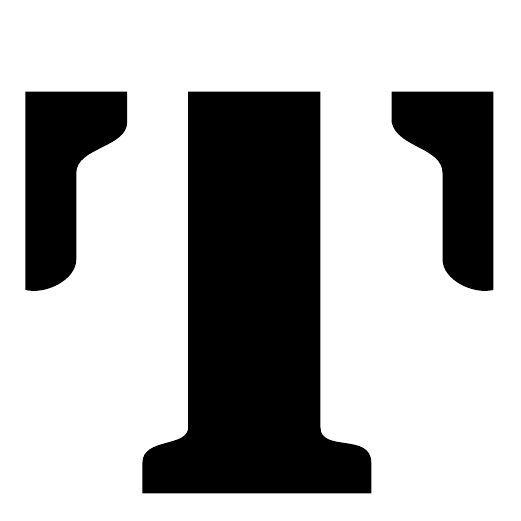} 
&
 \includegraphics[height=0.05\textwidth,width=0.05\textwidth]{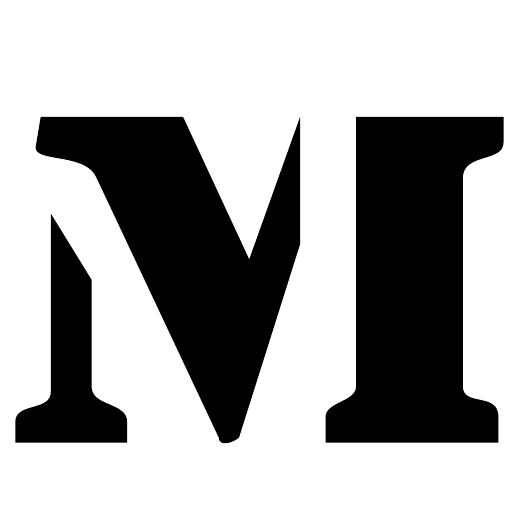} 
&
 \includegraphics[height=0.05\textwidth,width=0.05\textwidth]{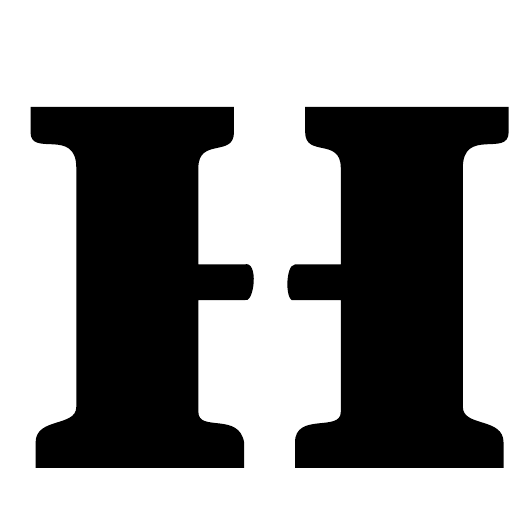} 
&
 \includegraphics[height=0.05\textwidth,width=0.05\textwidth]{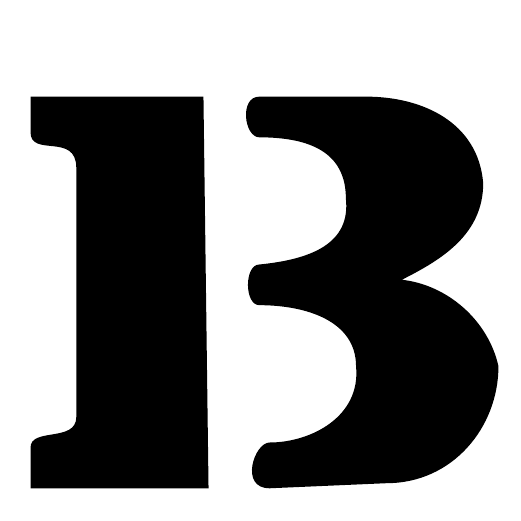} 
&
 \includegraphics[height=0.05\textwidth,width=0.05\textwidth]{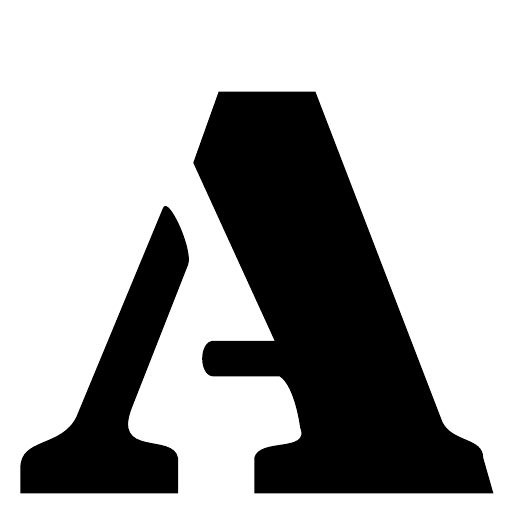} 
&
 \includegraphics[height=0.05\textwidth,width=0.05\textwidth]{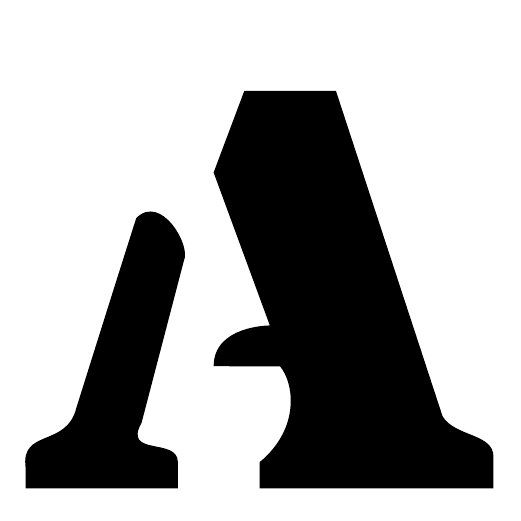} 
&
 \includegraphics[height=0.05\textwidth,width=0.05\textwidth]{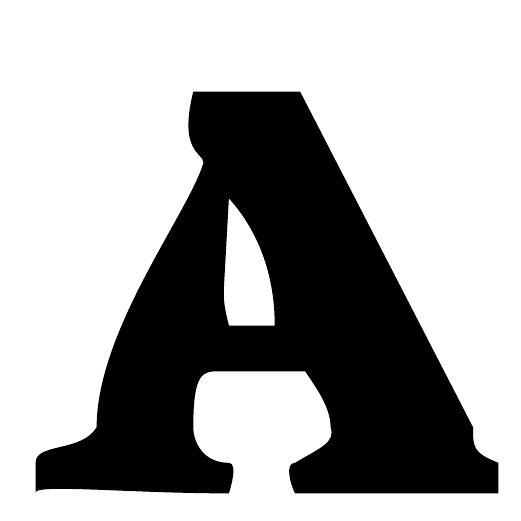}   
 &
 \includegraphics[height=0.06\textwidth,width=0.06\textwidth]{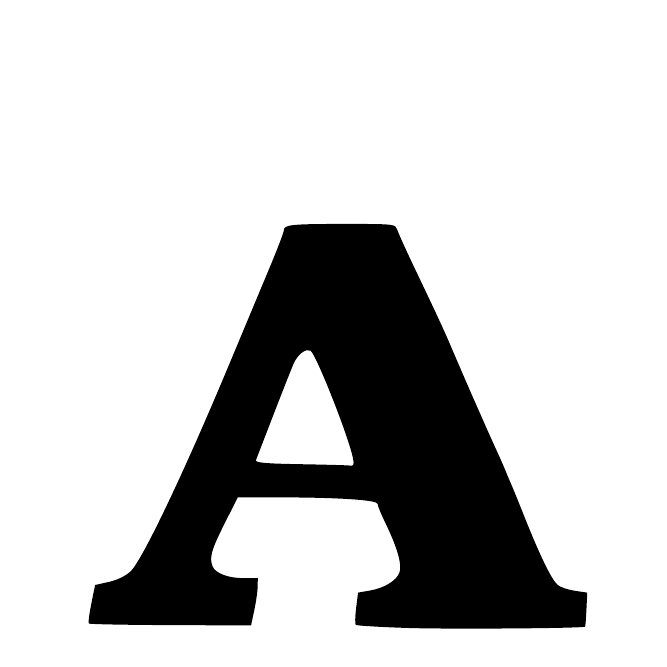} \vspace{-3mm} \\

 \includegraphics[height=0.055\textwidth,width=0.055\textwidth]{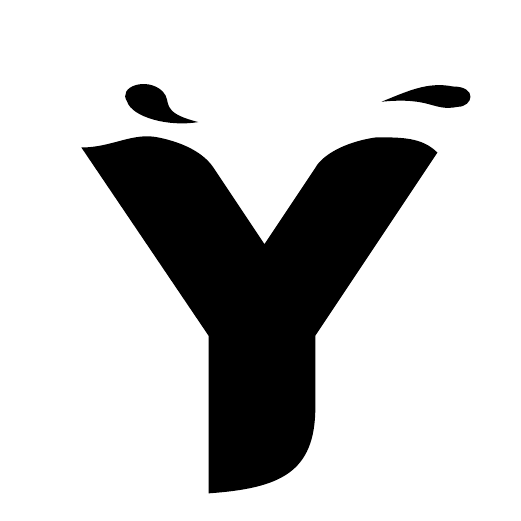} 
&
 \includegraphics[height=0.055\textwidth,width=0.055\textwidth]{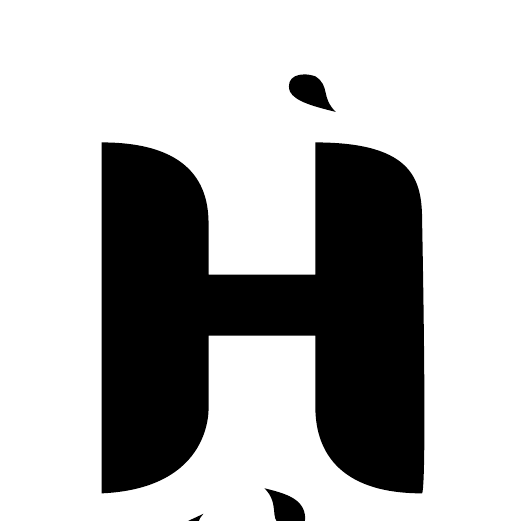} 
&
 \includegraphics[height=0.055\textwidth,width=0.055\textwidth]{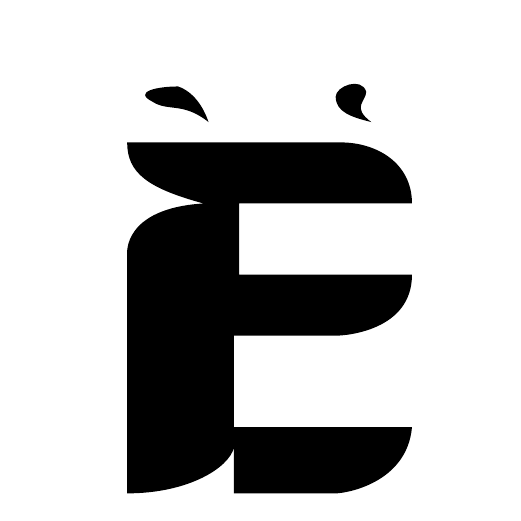} 
&
 \includegraphics[height=0.055\textwidth,width=0.055\textwidth]{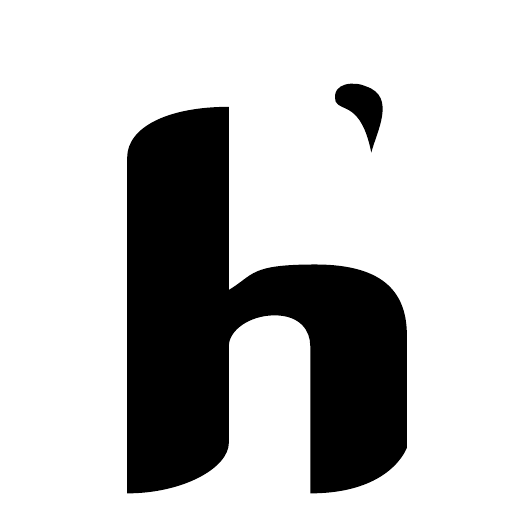} 
&
 \includegraphics[height=0.055\textwidth,width=0.055\textwidth]{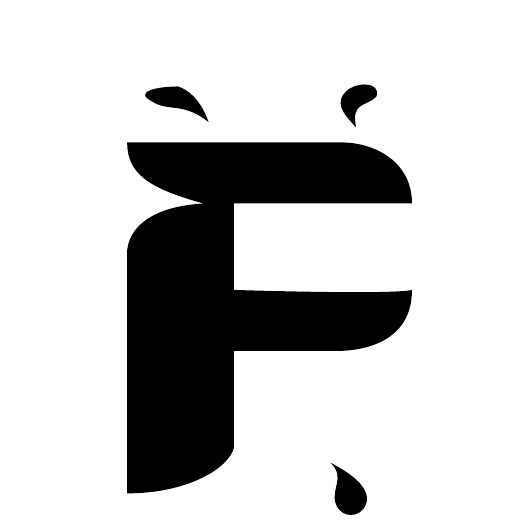} 
&
\includegraphics[height=0.055\textwidth,width=0.055\textwidth]{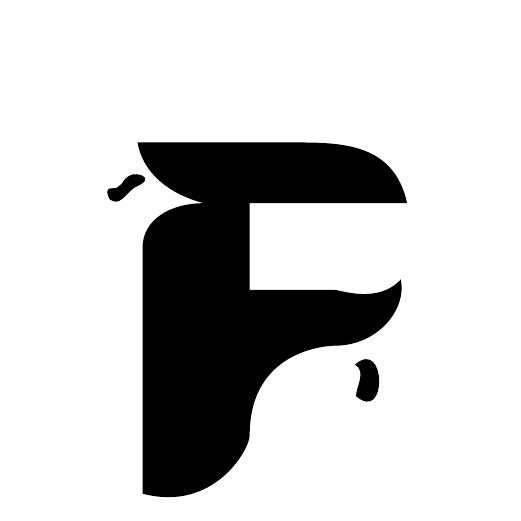} 
&
 \includegraphics[height=0.055\textwidth,width=0.055\textwidth]{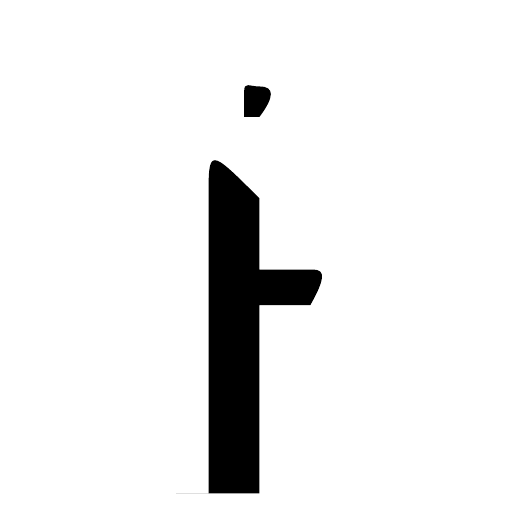}  
 &
 \includegraphics[height=0.065\textwidth,width=0.065\textwidth]{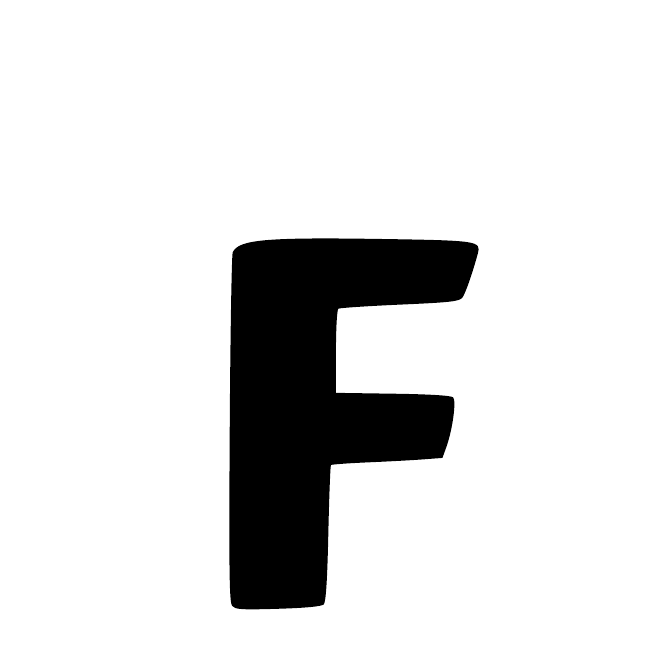} \vspace{-3mm} \\
 \includegraphics[height=0.055\textwidth,width=0.055\textwidth]{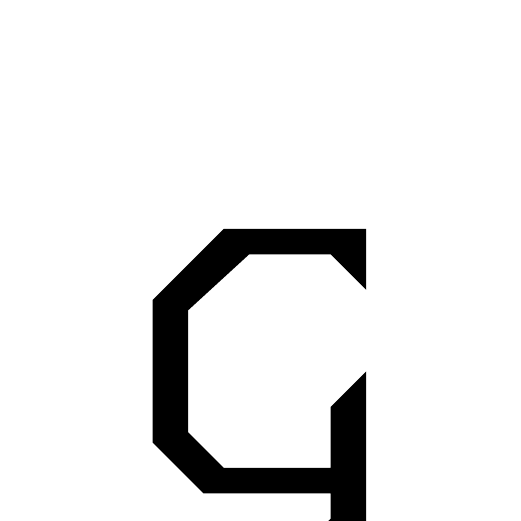} 
&
 \includegraphics[height=0.05\textwidth,width=0.05\textwidth]{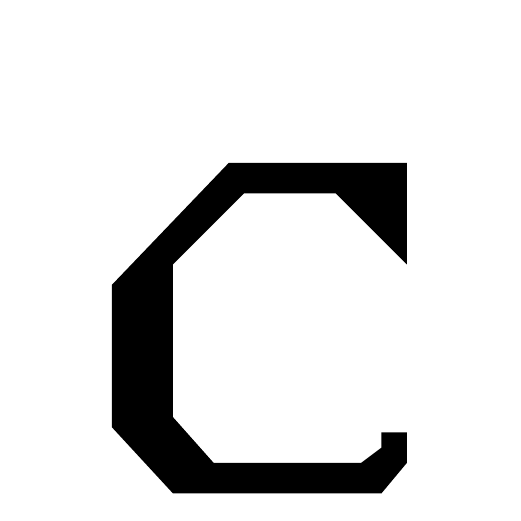} 
&
 \includegraphics[height=0.055\textwidth,width=0.055\textwidth]{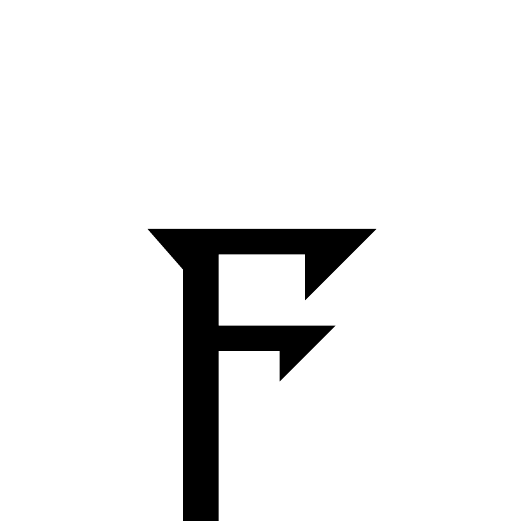} 
&
 \includegraphics[height=0.055\textwidth,width=0.055\textwidth]{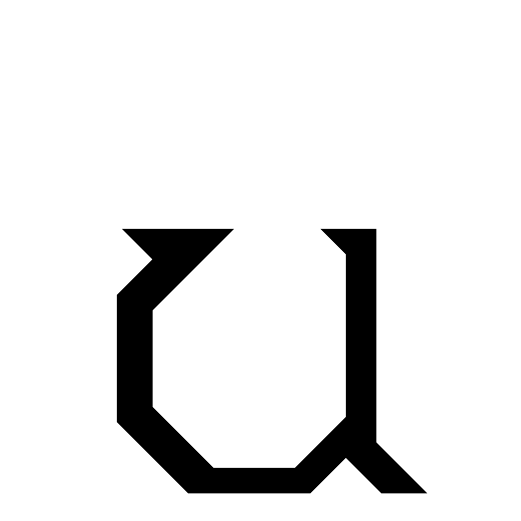} 
&
 \includegraphics[height=0.055\textwidth,width=0.055\textwidth]{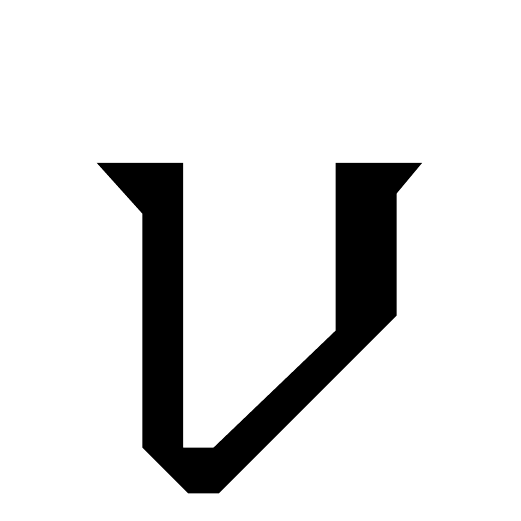} 
&
 \includegraphics[height=0.055\textwidth,width=0.055\textwidth]{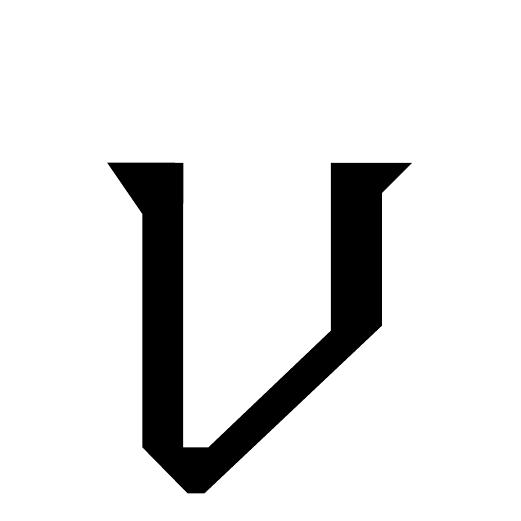} 
&
 \includegraphics[height=0.055\textwidth,width=0.055\textwidth]{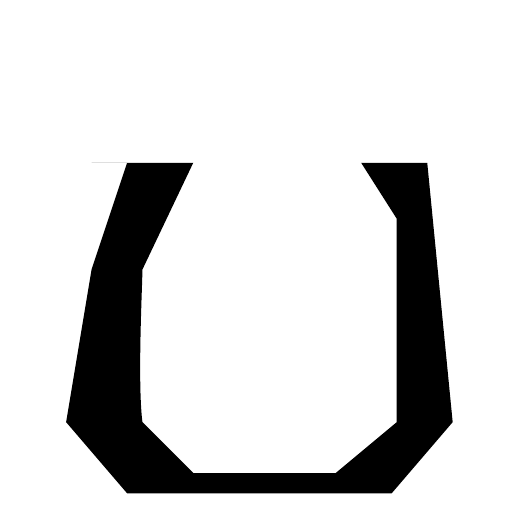}   
 &
 \includegraphics[height=0.055\textwidth,width=0.055\textwidth]{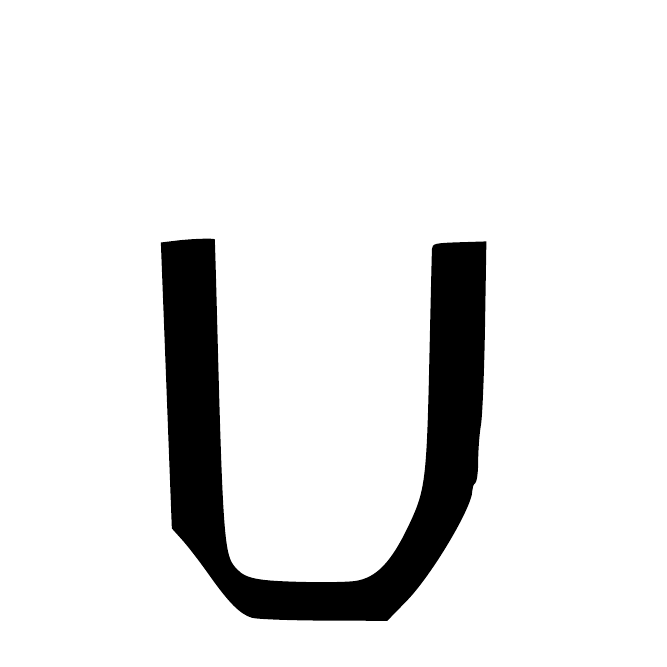} \\
 
    \end{tabular}
    \vspace{-4mm}
    \caption{Few-shot style transfer results. \emph{Left:} reference glyphs from a test font style. \emph{Right:} (a) artist-made (``ground-truth'') glyphs, (b) Ours, (c) DeepVecFont-v2 \cite{wang2023deepvecfont}, 
    and (d) DualVector \cite{dualvector}. 
    \label{fig:rebuttal_font_style_transfer}}
    \vspace{-1em}
\end{figure}

\begin{figure*}[t!]
    \centering
    \setlength{\tabcolsep}{1pt}
    \renewcommand{\arraystretch}{1}
    \vspace{-5mm}\newcommand{\includeimg}[1]{\includegraphics[width=0.9\linewidth]{#1}}

    % }
    \setlength{\arrayrulewidth}{0.001pt}

    \begin{tabular}{c}
        \includegraphics[width=.97\textwidth]{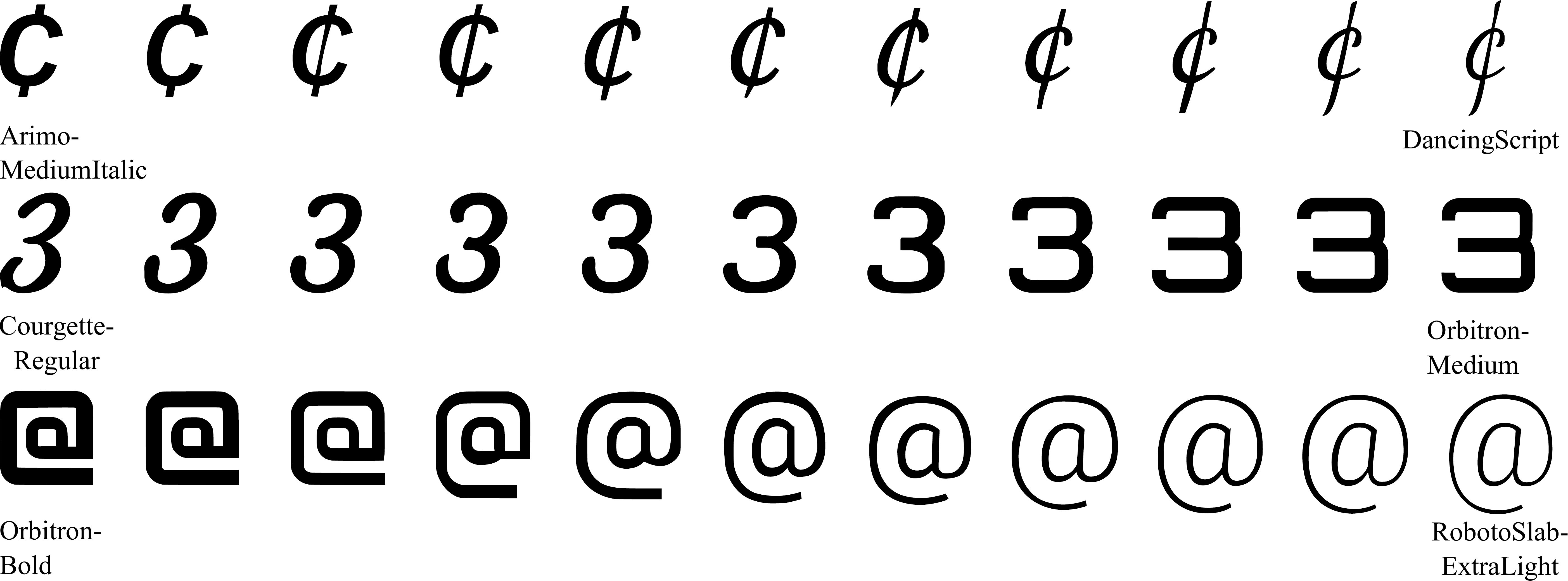}
    \end{tabular}
    \vspace{-3mm}
    \caption{Font interpolation. We perform linear interpolation in embedding space from source font (left) $\rightarrow$ target font (right).
    \vspace{-3mm}
    }
    \label{fig:font_interpolation}
    \vspace{-0em}
\end{figure*}

\vspace{-0.5em}
\paragraph{Qualitative Results.}
Figure \ref{fig:missing_testset_comp} shows qualitative comparisons for missing glyph generation. Compared to our method, we observe that ChiroDiff produces imprecise control points and curve structure, resulting in significant distortions and artifacts in the synthesized glyphs. We also observed degraded results in all alternative variants of our method in terms of misplaced or skipped control points. Additional results are also in the supplementary. % supplementary
Figures \ref{fig:teaser},\ref{fig:more_results} show additional results of missing glyph generation for our method on the Google Font dataset for various target fonts and characters. 
\rev{Figure \ref{fig:stochasticity} demonstrates multiple samples generated by our diffusion pipeline with random seeds. The samples adhere to the same font style, while having subtle variation in the glyph proportions and control point distributions. From a practical point of view, a designer can explore multiple such samples, and choose the most preferred variant.} 

% In Figure \ref{fig:vec_compare}, 
In the supplementary, instead of using the vector diffusion model, we use off-the-shelf vectorization methods on the raster glyph image produced by the raster diffusion stage. As shown in the comparison, this approach often fails to produce coherent curve topology and structure.

\subsection{Few-shot font style transfer}
\label{sec:font-style-transfer}

For this application, we compare with DeepVecFont-v2 \cite{wang2023deepvecfont} and DualVector \cite{dualvector}, both of which previously demonstrated few-shot font style transfer. To perform the comparison, we use the dataset proposed in the DeepVecFont-v2 paper. The dataset includes $52$ lowercase and uppercase Latin characters in various font styles -- there are total $8,035$ training fonts and $1,425$ test ones. Each test case contains $4$ reference characters from a novel font (unobserved during training). The reference characters are available in both vector and raster format.
 Methods are supposed to transfer this novel font style to testing characters. 
 We note that DeepVecFont-v2 requires the vector representation of the reference characters as additional input condition, while DualVector and our method only use the raster reference images.

\vspace{-0.6em}
\paragraph{Quantitative results.} 
\rev{Table \ref{tab:font_style_transfer_rebuttal} shows numerical comparisons based on the same evaluation metrics as in the missing glyph generation application. 
Our method outperforms DeepVecFont-v2 and DualVector on all metrics.}

\vspace{-0.6em}
\paragraph{Qualitative results.} Figure \ref{fig:rebuttal_font_style_transfer} demonstrates font style transfer results. We observed that both \rev{DeepVecFont-v2 tends to succeed in capturing the font style of the reference glyphs, yet still often produces subtle distortions in the vector paths. Our method produces results that match the style of references glyphs with less artifacts, even in challenging styles. Additional results are in Figure \ref{fig:teaser} and in the supplementary.
% \ref{fig:font_style_transfer_supp}.
}

\subsection{Font style interpolation}
\label{sec:interpolation}

Finally, we experimented with interpolating two given font styles.
To perform interpolation, we first obtain the font emdeddings from our trained look-up table, then perform linear interpolation of these embeddings. Our diffusion model is then conditioned on the interpolated embedding vector for font style. We demonstrate qualitative results in Figure \ref{fig:font_interpolation}. Our results smoothly interpolates artistic properties of the source and target font, such as the variable stroke width and local curvature, while preserving structural and topological properties of the glyphs e.g., their genus.

\section{Conclusion}
\label{sec:conclusion}
We presented a generative model of vector fonts. 
\rev{We show that a cascade of a raster and vector diffusion model can overcome the challenges of neural parametric curve prediction, and generate editable vector fonts with precise geometry and control point locations.}

\begin{wrapfigure}{R}{0.45\columnwidth}
  \vspace{-3mm} 
  \includegraphics[width=0.45\columnwidth]{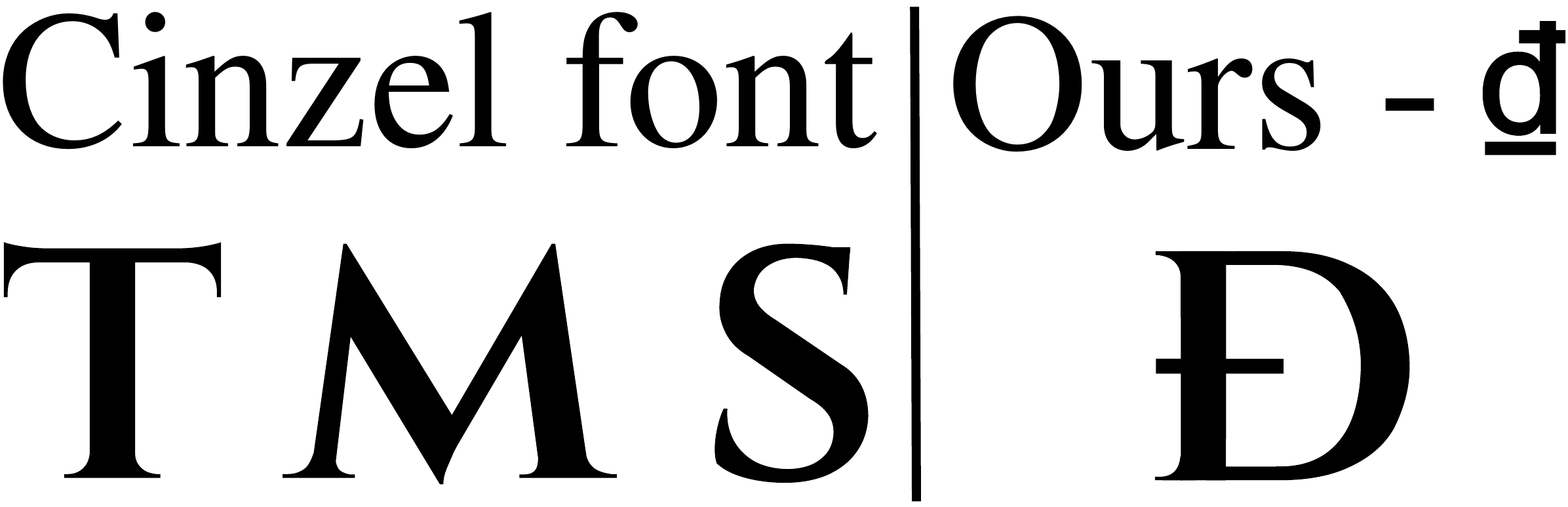}
  \vspace{-6mm} 
  \caption{Failure case. }
  \vspace{-2mm}
\label{fig:limitation}
\end{wrapfigure}

\vspace{-1em}
\paragraph{Limitations.} 
\rev{We require vector paths as supervision and cannot take advantage of raster images as additional supervision that may be available.
Another limitation is shown in Figure \ref{fig:limitation}. Cinzel is an uppercase font. For the missing glyph \textdong (dong), a lowercase glyph, our method ``hallucinates''  an uppercase glyph. 
While the generated glyph preserves the style of the font and the structure of the glyph (the stroke), it does not preserve the symbolic meaning of the glyph.} 

\vspace{-1em}
\paragraph{Future work.} Generative models of  images in the raster domain have been successfully used to learn complex priors about the visual world that can be used in various tasks such as inpainting and 3D tasks. 
We believe that a generative model for vector graphics in general could have several downstream applications, such as automatic generation and completion of logos, icons, line drawings and illustrations \cite{Kalogerakis:2009:ddcurvature,Liu_2020_CVPR,Liu_2021_ICCV}, or be extended to produce 3D parametric curves and surfaces \cite{sharma2020parsenet}.

\vspace{-1em}
\paragraph{Acknowledgments.} Our project was funded by Adobe Research.

%%%%%%%%% REFERENCES
{\small
\bibliographystyle{ieee_fullname}
\bibliography{main}
}

\newpage

% \title{Appendix}
% \author{}
% \date{\vspace{-5ex}}
% \maketitle

{\Large \textbf{Appendix}}

\section*{A: Implementation Details}
\label{sec:implementation_details}

Here we provide additional implementation details of our network architecture. Our model is implemented in PyTorch.

\paragraph{Raster diffusion denoiser.} Our raster diffusion denoiser follows the UNet architecture in \cite{dhariwal2021diffusion, rombach2022high}. The UNet model uses a stack of residual layers and downsampling
convolutions, followed by a stack of residual layers with upsampling convolutions, with skip connections connecting the layers with the same spatial size. We provide an overview of the hyperparameters in Table \ref{table:raster_arch}. 
To condition the model on the character identifier, we use a look-up table to project it to an $896$-dimensional embedding and then add it together with  the time step embedding to modulate the feature maps of each residual block.

\paragraph{Few-shot font style encoder.} 
In the application of few-shot font style transfer, we used a ConvNet to encode reference raster glyphs into a font style feature map. We used the encoder part of the UNet architecture in \cite{dhariwal2021diffusion, rombach2022high}. The ConvNet encoder encodes the $64\times 64$ input image into an $8 \times 8 \times 512$ high-dimensional feature map via $3$ downsampling layers.

\paragraph{Vector diffusion denoiser.} 
Our vector diffusion denoiser is an encoder-only transformer following BERT \cite{devlin2018bert}. We set the number of transformer layers and the number of attention heads to $8$ and $12$ respectively.
To condition the vector diffusion denoiser on the raster guidance $x_0$, we first encode the $64 \times 64 \times 4$ raster image to a $16 \times 16 \times 768$ high-dimensional feature map with a ConvNet encoder. The ConvNet encoder has two downsampling layers with self-attention layers at resolution $32\times 32$ and $16\times 16$. The ConvNet encoder is trained with the transformer jointly.
After obtaining the $16 \times 16 \times 768$ high-dimensional feature map, we flatten it to a shape of $256 \times 768$, then we add it to each transformer layer via cross-attention following \cite{rombach2022high}.

\paragraph{Computation cost.}  Raster-DM and Vector-DM are trained separately. Each of them is trained on 8 A100 GPUs for 5 days. Finally, at inference time, generating a glyph takes around 10 seconds on a A100.

\begin{table}[tbp]
\renewcommand{\tabcolsep}{4.5pt}  % horizontal column padding
\centering
\begin{tabular}{lc}
\toprule
Input shape & 64 $\times$ 64 $\times$ 4\\
Diffusion steps & 1000 \\
Noise Schedule & cosine \\
Channels & 224\\
Depth & 2\\
Channel Multiplier & 1,2,3,4\\
Attention resolutions & 32,16,8\\
Head Channels & 32\\
Batch Size & 448\\
Learning Rate & 3.24e-5\\
\bottomrule
\end{tabular}
\vspace{1mm}
\caption{Hyperparameters for raster diffusion denoiser}

\label{table:raster_arch}
\end{table}

\section*{B: Additional Results}

\begin{figure*}[t]
\vspace{-8mm}
    \centering
    \includegraphics[width=1.0\linewidth]{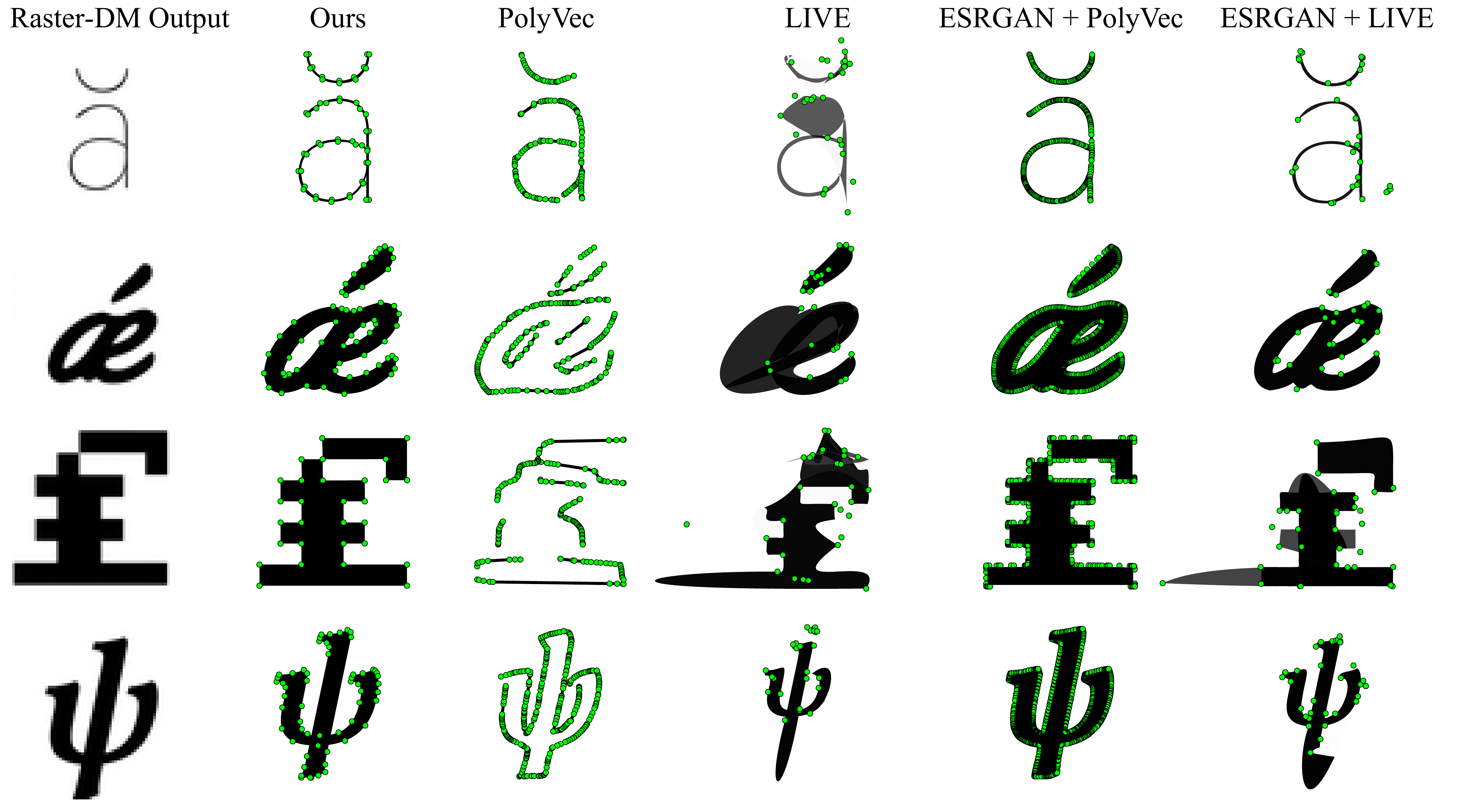}
    \vspace{-6mm}
    \caption{We compare our results (Ours) with PolyVec \cite{polyvec} and LIVE \cite{xu2022live} applied to the raster image produced by our raster diffusion stage (left-most column). We also compare with PolyVec and LIVE applied to a higher-resolution version of the raster image upsampled via ESRGAN \cite{wang2021real}.
    For each glyph, we show the predicted control points as well. Using our vector diffusion stage instead of an off-the-shelf vectorizer produces higher-quality glyphs and much more plausible control point distributions. Compared to our vector diffusion model, ESRGAN + PolyVec requires about ten times more control points for effective glyph reconstruction but sacrifices user editability and SVG compactness. We recommend the viewer to zoom in for better clarity.}
    \label{fig:vec_compare}
\end{figure*}

\paragraph{Comparison with a vectorizer approach.} 
As an alternative comparison, we tried the following approach: instead of using the vector diffusion model, we use PolyVec \cite{polyvec} or LIVE \cite{xu2022live} on the rasterized font image produced by our raster diffusion stage. We also tried upsampling the $64\times64$ output raster image to $256\times256$ using ESRGAN \cite{wang2021real} before passing it to the vectorizer. 
We show qualitative comparison in Figure \ref{fig:vec_compare}. In both cases, PolyVec and LIVE often failed to produce coherent curve topology,  structure, and plausible control point distributions. 

\begin{figure}[h]
    \centering
    \setlength{\tabcolsep}{2pt}
    \renewcommand{\arraystretch}{2}

    \begin{tabular}{ll  c c : c c}
        \multicolumn{2}{c}{Input reference} &  GT &  NNs & Ours & DVF-v2\\
        % \hline
        % &  & &  \\
\includegraphics[height=0.055\textwidth,width=0.055\textwidth]{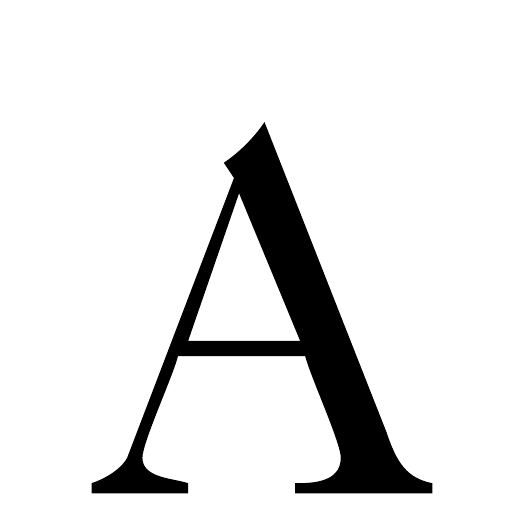} 
&
 \includegraphics[height=0.055\textwidth,width=0.055\textwidth]{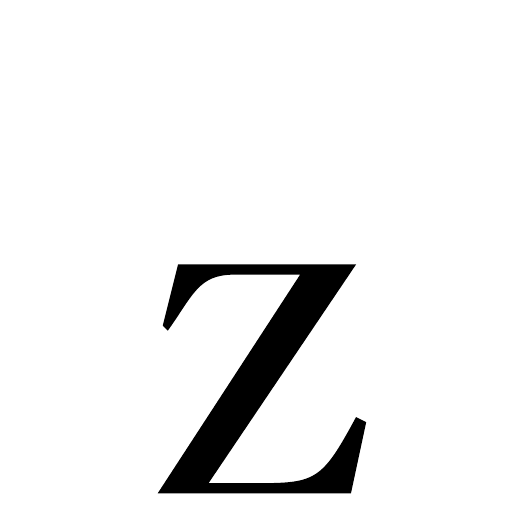} 
&
%  \includegraphics[height=0.055\textwidth,width=0.055\textwidth]{images/few_shot_gen/CVPR/pdfs/2697_ref2.pdf} 
% &
%  \includegraphics[height=0.055\textwidth,width=0.055\textwidth]{images/few_shot_gen/CVPR/pdfs/2697_ref3.pdf} 
% &
 \includegraphics[height=0.055\textwidth,width=0.055\textwidth]{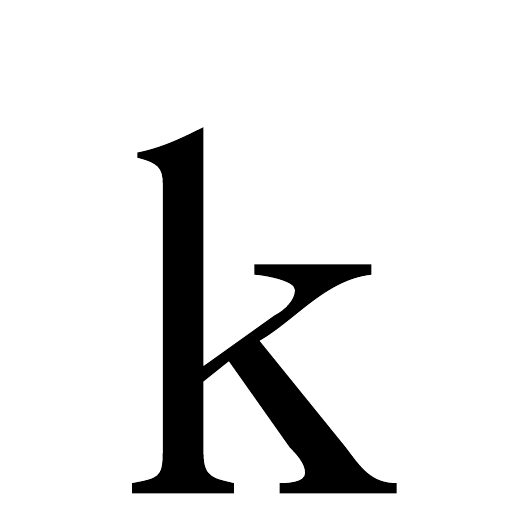} 
&
 \includegraphics[height=0.055\textwidth,width=0.055\textwidth]{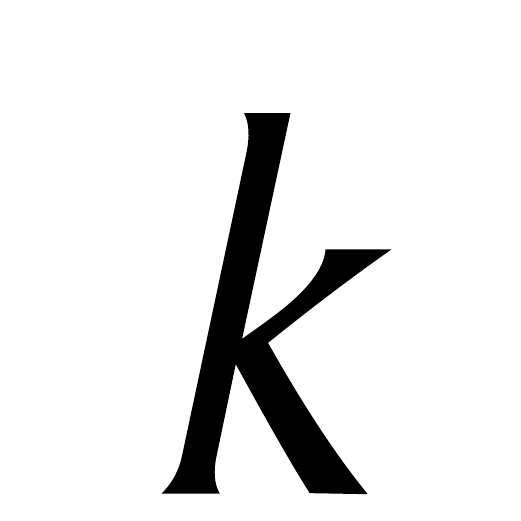}  
 &
 \includegraphics[height=0.055\textwidth,width=0.055\textwidth]{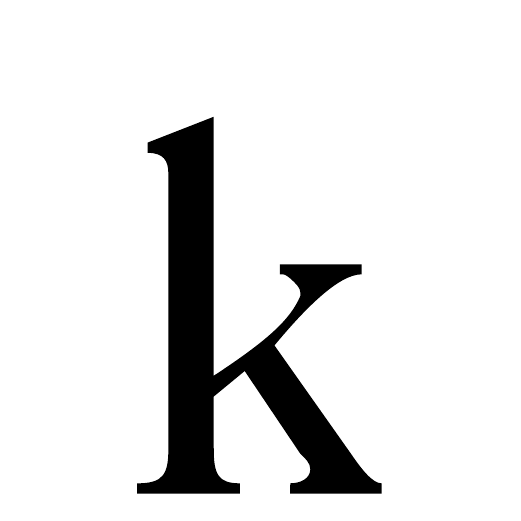} 
 &
 \includegraphics[height=0.055\textwidth,width=0.055\textwidth]{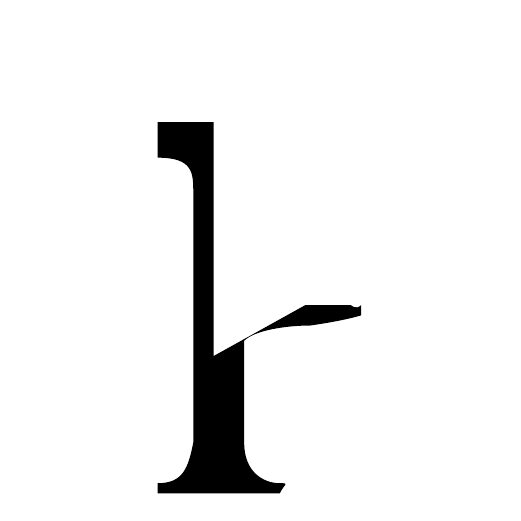} \\
\includegraphics[height=0.055\textwidth,width=0.055\textwidth]{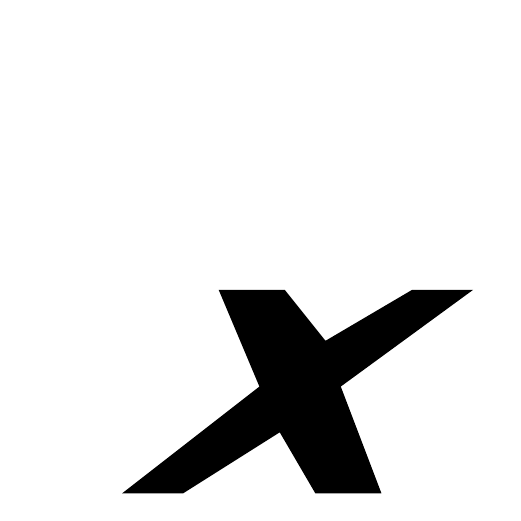} 
&
 \includegraphics[height=0.055\textwidth,width=0.055\textwidth]{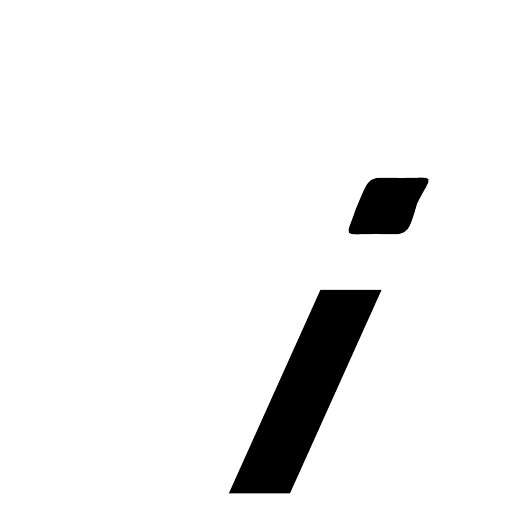} 
&
%  \includegraphics[height=0.045\textwidth,width=0.045\textwidth]{images/few_shot_gen/CVPR/pdfs/3_ref2.pdf} 
% &
%  \includegraphics[height=0.045\textwidth,width=0.045\textwidth]{images/few_shot_gen/CVPR/pdfs/3_ref3.pdf} 
% &
 \includegraphics[height=0.055\textwidth,width=0.055\textwidth]{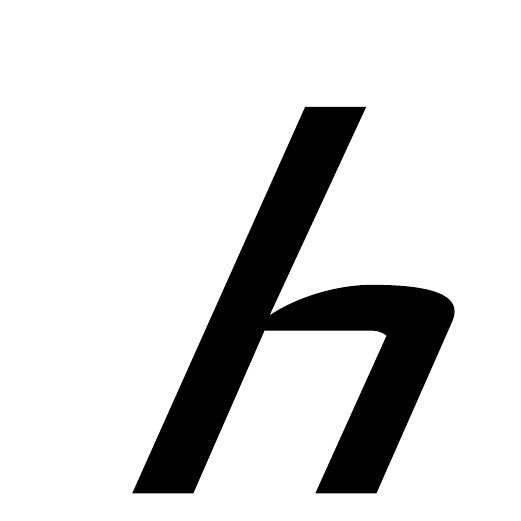} 
&
 \includegraphics[height=0.055\textwidth,width=0.055\textwidth]{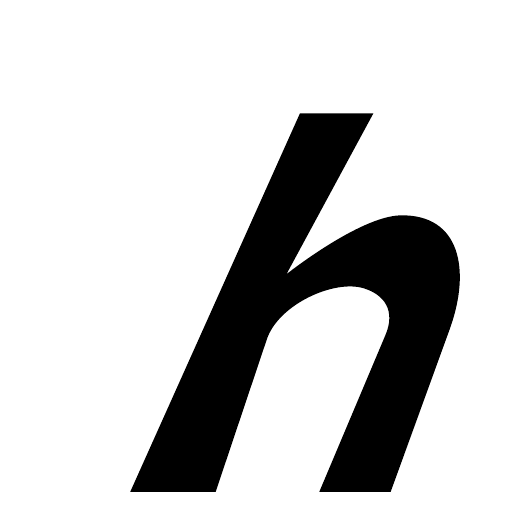} 
 &
 \includegraphics[height=0.055\textwidth,width=0.055\textwidth]{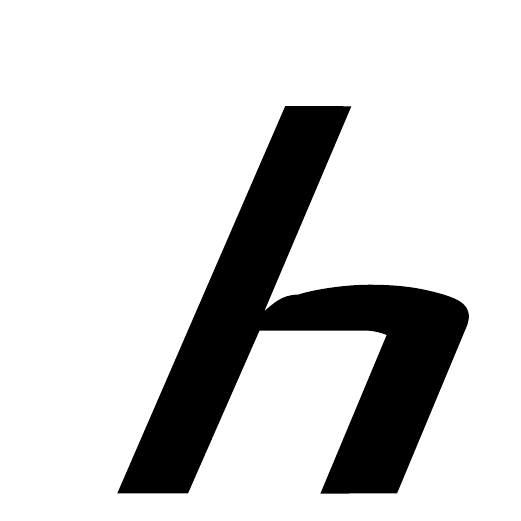} 
 &
 \includegraphics[height=0.055\textwidth,width=0.055\textwidth]{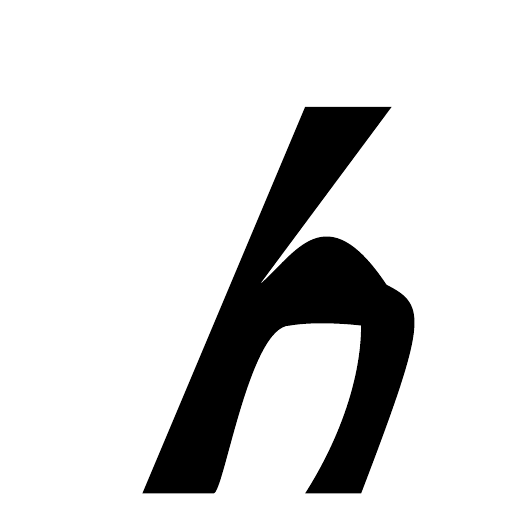}\\
\includegraphics[height=0.055\textwidth,width=0.055\textwidth]{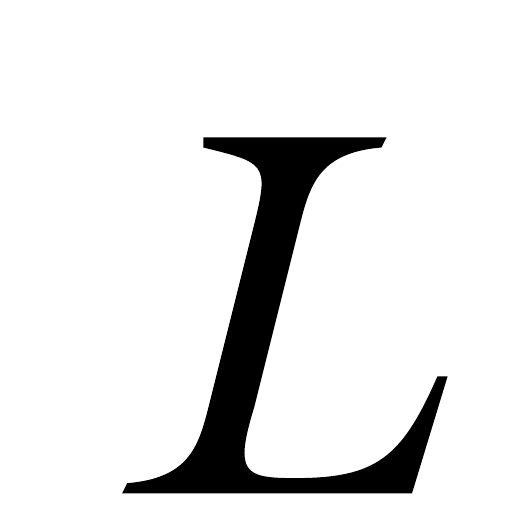} 
&
 \includegraphics[height=0.055\textwidth,width=0.055\textwidth]{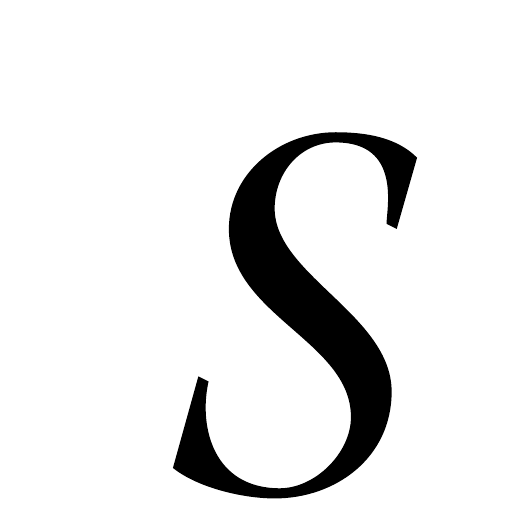} 
&
%  \includegraphics[height=0.055\textwidth,width=0.055\textwidth]{images/few_shot_gen/CVPR/pdfs/3148_ref2.pdf} 
% &
%  \includegraphics[height=0.055\textwidth,width=0.055\textwidth]{images/few_shot_gen/CVPR/pdfs/3148_ref3.pdf} 
% &
 \includegraphics[height=0.055\textwidth,width=0.055\textwidth]{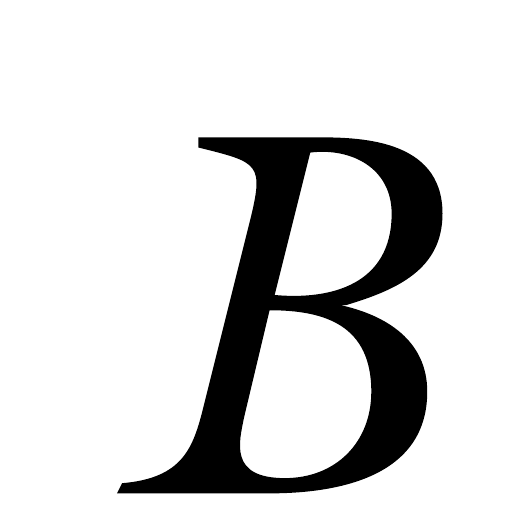} 
&
 \includegraphics[height=0.055\textwidth,width=0.055\textwidth]{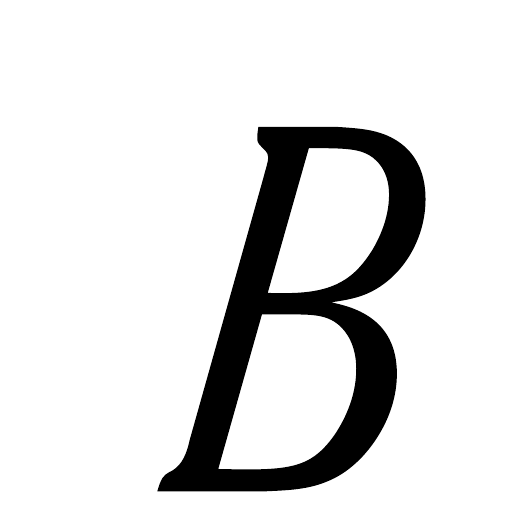} 
 &
 \includegraphics[height=0.055\textwidth,width=0.055\textwidth]{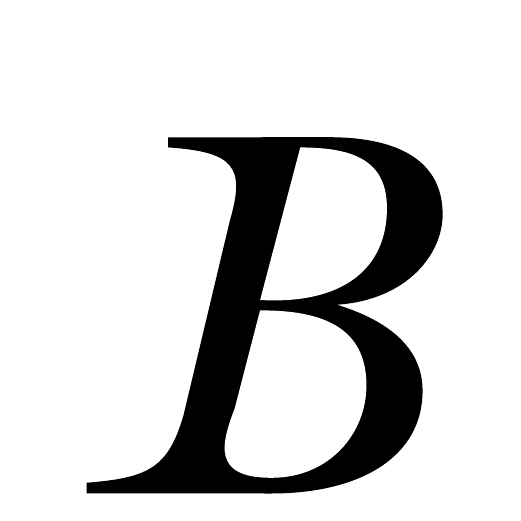} 
 &
 \includegraphics[height=0.055\textwidth,width=0.055\textwidth]{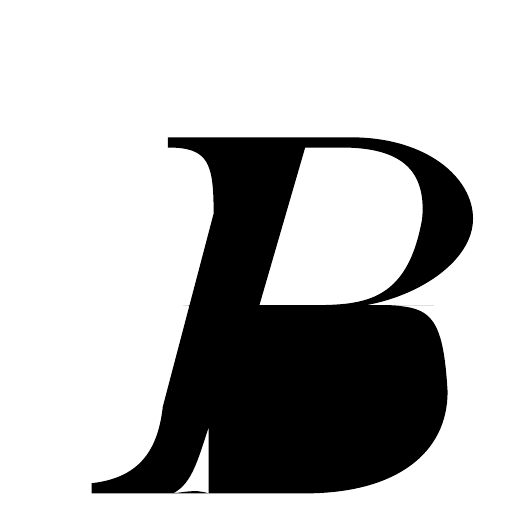} \\
\includegraphics[height=0.055\textwidth,width=0.055\textwidth]{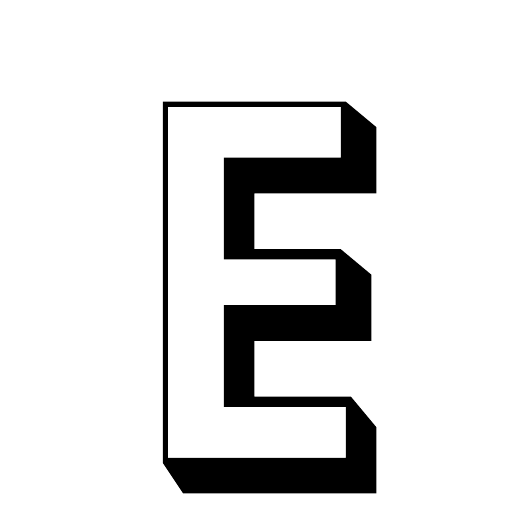} 
&
 \includegraphics[height=0.055\textwidth,width=0.055\textwidth]{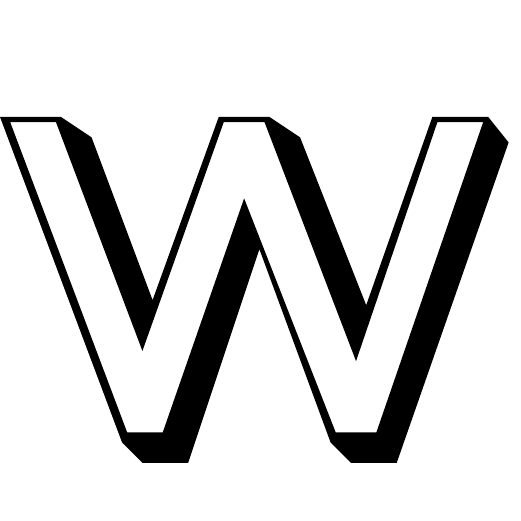} 
&
%  \includegraphics[height=0.055\textwidth,width=0.055\textwidth]{images/few_shot_gen/CVPR/pdfs/34_ref2.pdf} 
% &
%  \includegraphics[height=0.055\textwidth,width=0.055\textwidth]{images/few_shot_gen/CVPR/pdfs/34_ref3.pdf} 
% &
 \includegraphics[height=0.055\textwidth,width=0.055\textwidth]{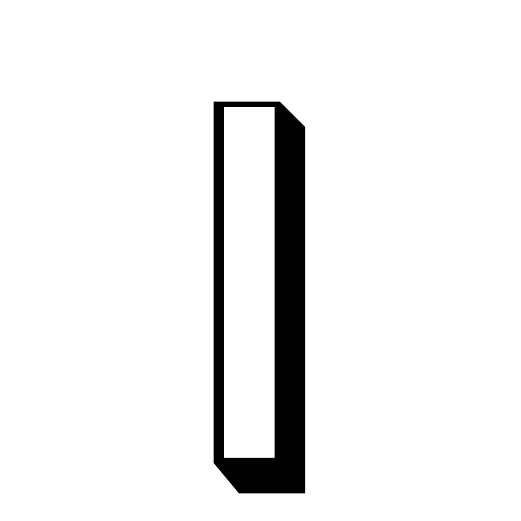} 
&
 \includegraphics[height=0.055\textwidth,width=0.055\textwidth]{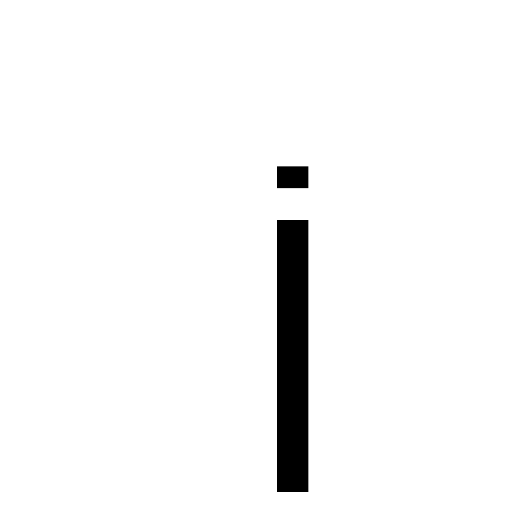}  
 &
  \includegraphics[height=0.055\textwidth,width=0.055\textwidth]{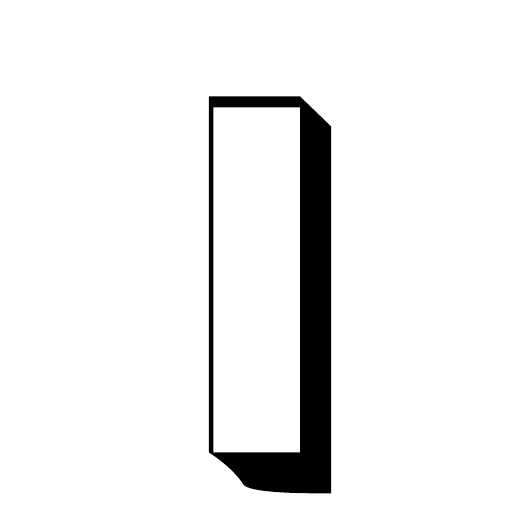} 
&
 \includegraphics[height=0.055\textwidth,width=0.055\textwidth]{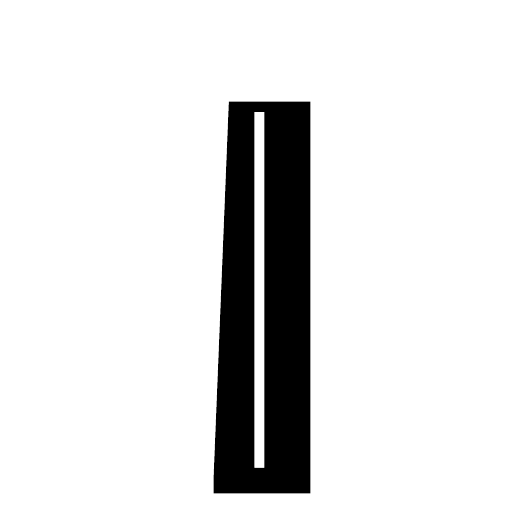} \\
\includegraphics[height=0.055\textwidth,width=0.055\textwidth]{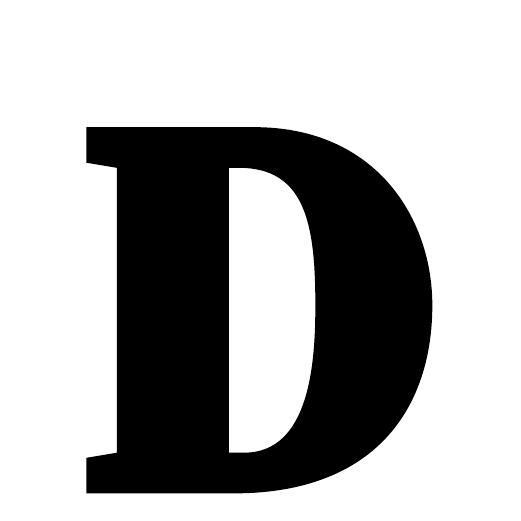} 
&
 \includegraphics[height=0.055\textwidth,width=0.055\textwidth]{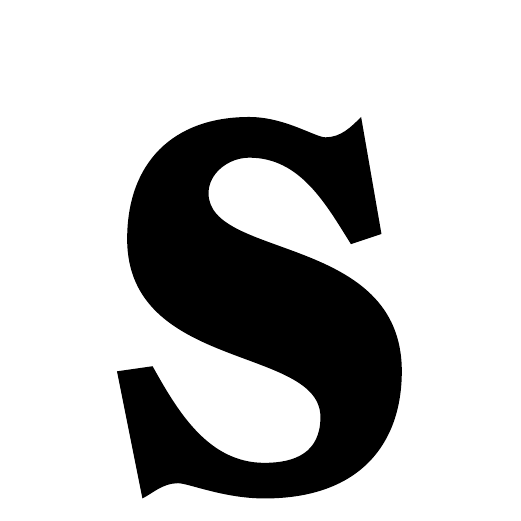} 
&
%  \includegraphics[height=0.055\textwidth,width=0.055\textwidth]{images/few_shot_gen/CVPR/pdfs/6772_ref2.pdf} 
% &
%  \includegraphics[height=0.055\textwidth,width=0.055\textwidth]{images/few_shot_gen/CVPR/pdfs/6772_ref3.pdf} 
% &
 \includegraphics[height=0.055\textwidth,width=0.055\textwidth]{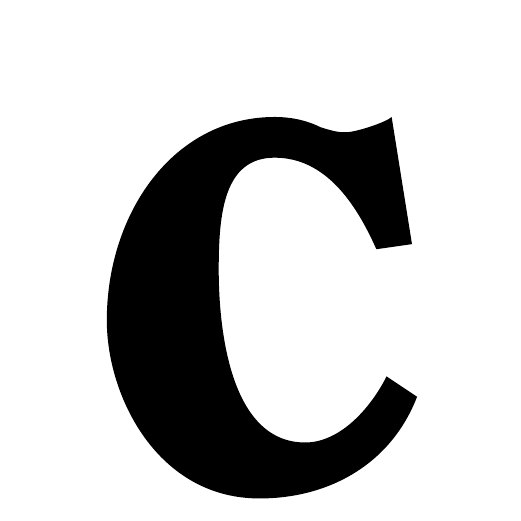} 
&
 \includegraphics[height=0.055\textwidth,width=0.055\textwidth]{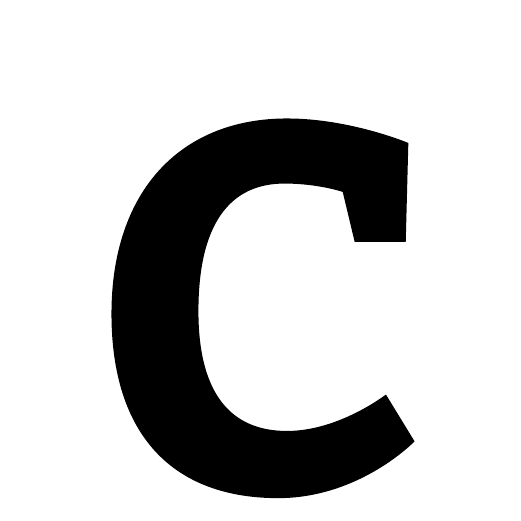}
 &
 \includegraphics[height=0.055\textwidth,width=0.055\textwidth]{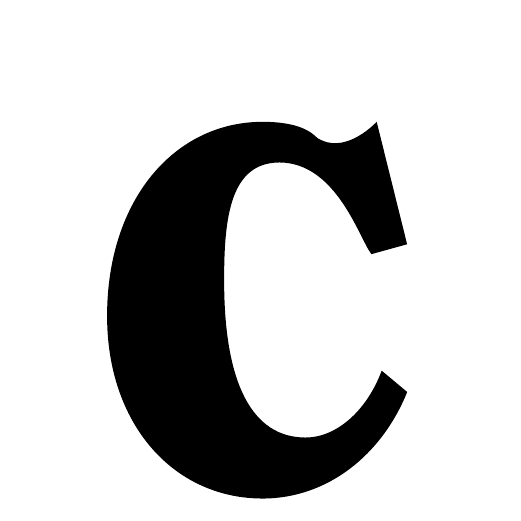} 
 &
 \includegraphics[height=0.055\textwidth,width=0.055\textwidth]{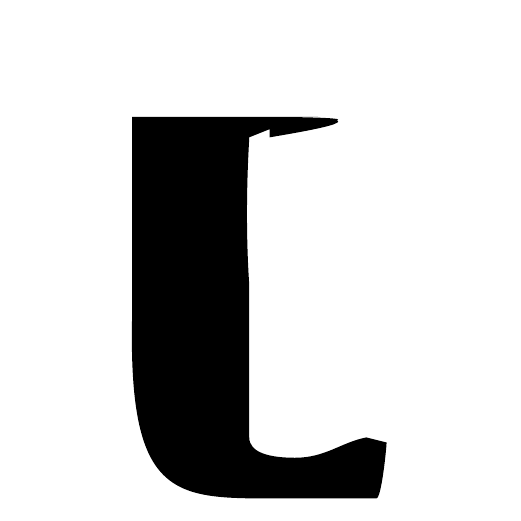} \\
\includegraphics[height=0.05\textwidth,width=0.05\textwidth]{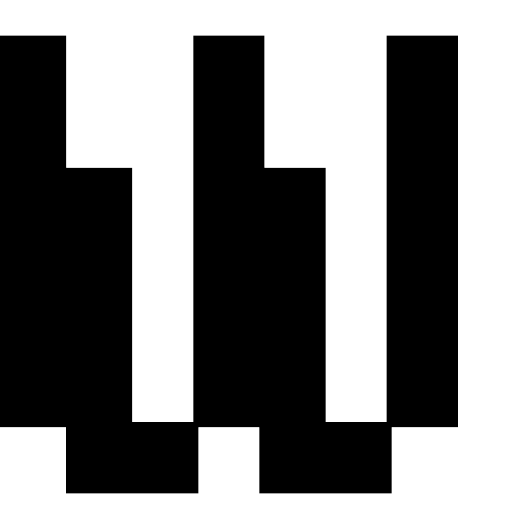} 
&
 \includegraphics[height=0.05\textwidth,width=0.05\textwidth]{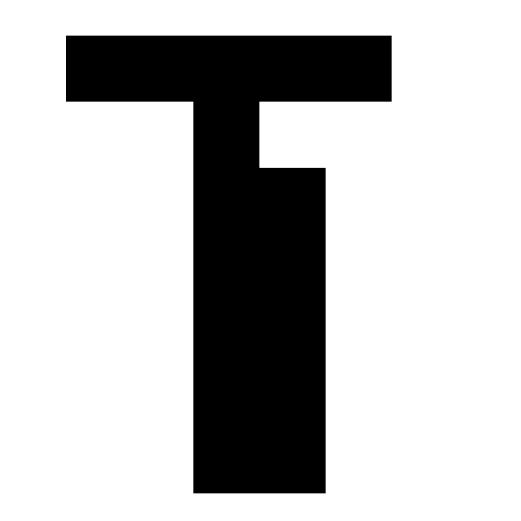} 
&
%  \includegraphics[height=0.05\textwidth,width=0.05\textwidth]{images/few_shot_gen/CVPR/pdfs/89_ref2.pdf} 
% &
%  \includegraphics[height=0.05\textwidth,width=0.05\textwidth]{images/few_shot_gen/CVPR/pdfs/89_ref3.pdf} 
% &
 \includegraphics[height=0.05\textwidth,width=0.05\textwidth]{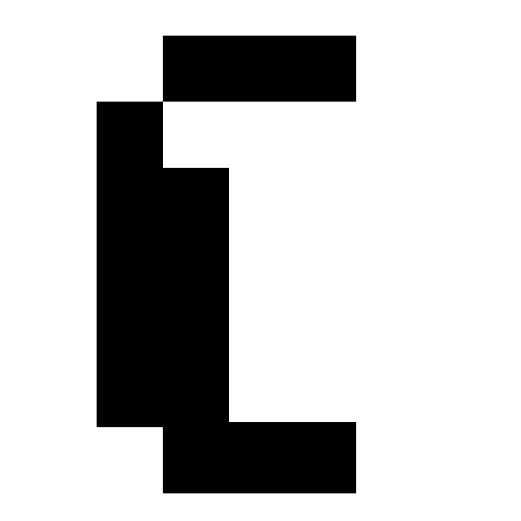} 
&
 \includegraphics[height=0.05\textwidth,width=0.05\textwidth]{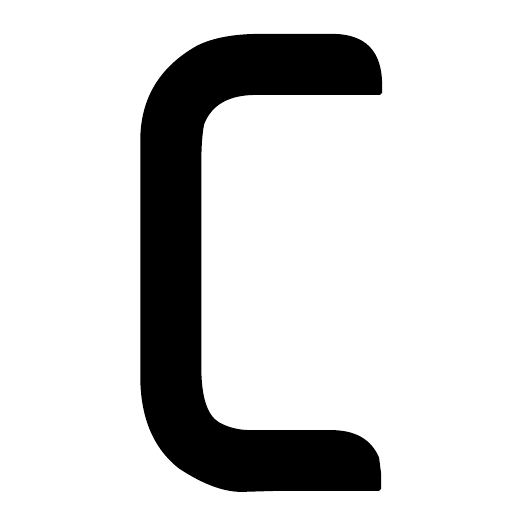}  
 &
 \includegraphics[height=0.05\textwidth,width=0.05\textwidth]{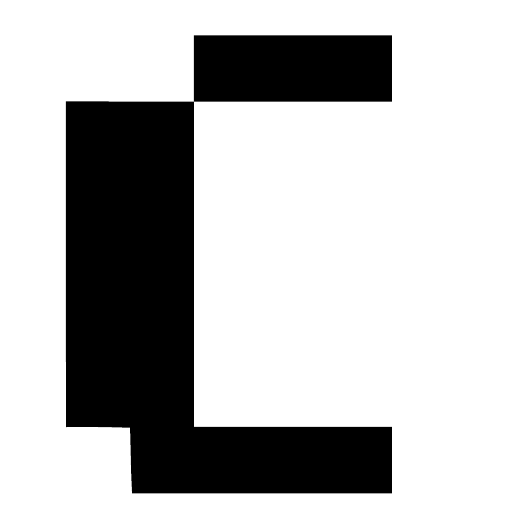} 
 &
 \includegraphics[height=0.05\textwidth,width=0.05\textwidth]{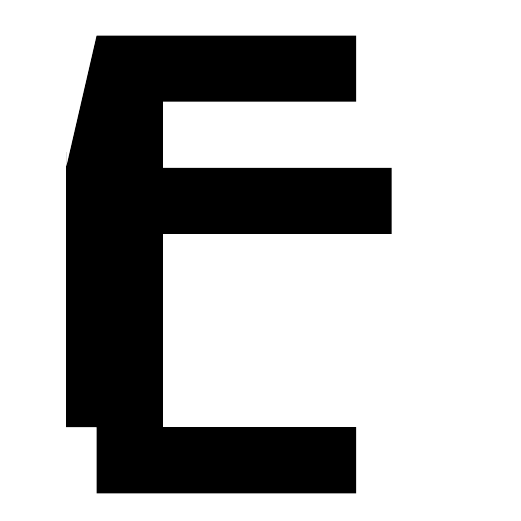} \\

 \includegraphics[height=0.05\textwidth,width=0.05\textwidth]{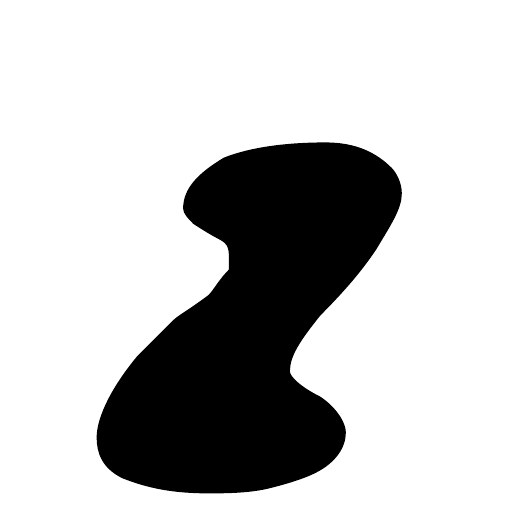} 
 &
  \includegraphics[height=0.05\textwidth,width=0.05\textwidth]{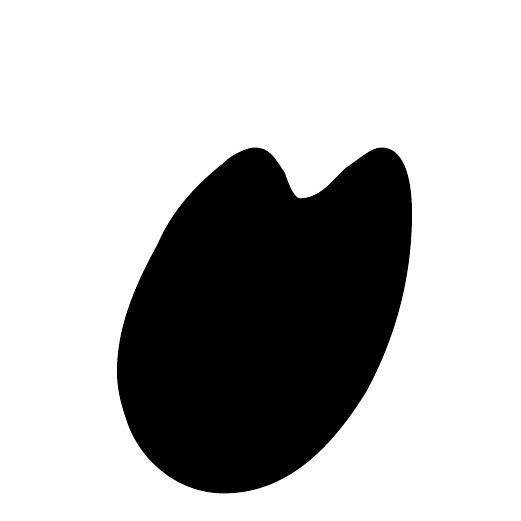} 
 &
 %  \includegraphics[height=0.05\textwidth,width=0.05\textwidth]{images/few_shot_gen/CVPR/pdfs/1562_ref2.pdf} 
 % &
 %  \includegraphics[height=0.05\textwidth,width=0.05\textwidth]{images/few_shot_gen/CVPR/pdfs/1562_ref3.pdf} 
 % &
  \includegraphics[height=0.05\textwidth,width=0.05\textwidth]{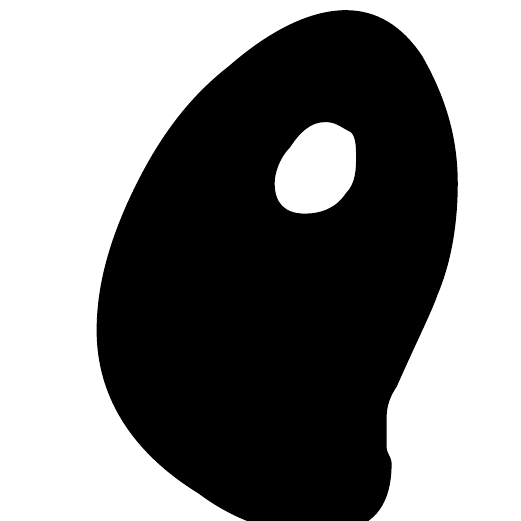} 
 &
  \includegraphics[height=0.05\textwidth,width=0.05\textwidth]{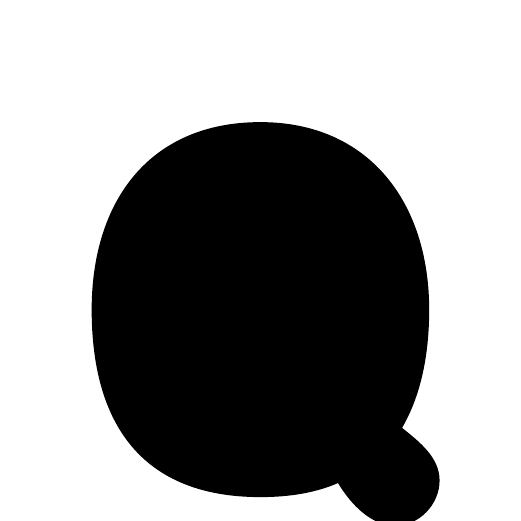} 
   &
  \includegraphics[height=0.05\textwidth,width=0.05\textwidth]{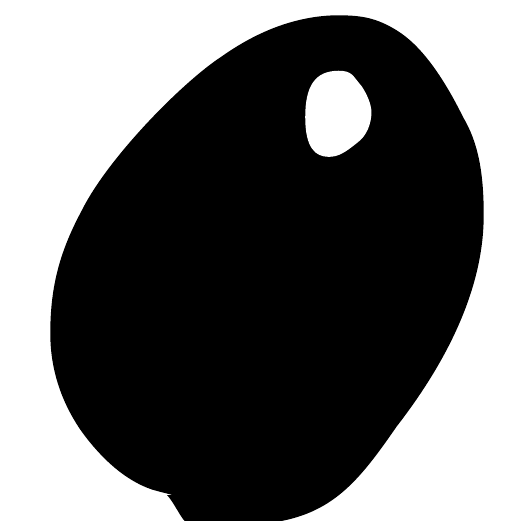} 
   &
  \includegraphics[height=0.05\textwidth,width=0.05\textwidth]{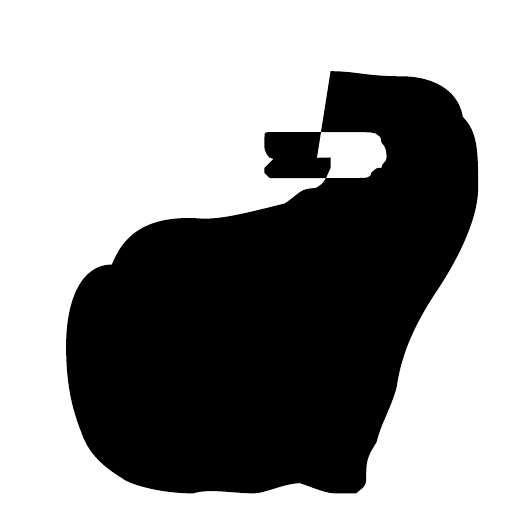} \\
\includegraphics[height=0.055\textwidth,width=0.055\textwidth]{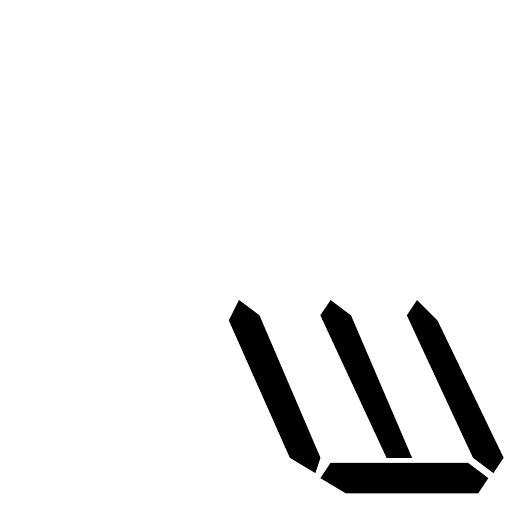} 
&
 \includegraphics[height=0.055\textwidth,width=0.055\textwidth]{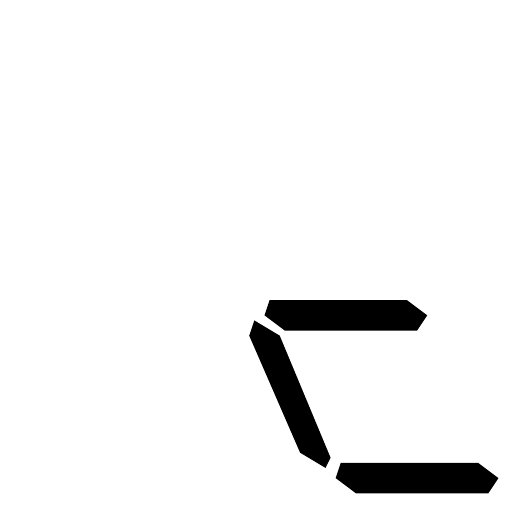} 
&
%  \includegraphics[height=0.055\textwidth,width=0.055\textwidth]{images/few_shot_gen/CVPR/pdfs/2024_ref2.pdf} 
% &
%  \includegraphics[height=0.055\textwidth,width=0.055\textwidth]{images/few_shot_gen/CVPR/pdfs/2024_ref3.pdf} 
% &
 \includegraphics[height=0.055\textwidth,width=0.055\textwidth]{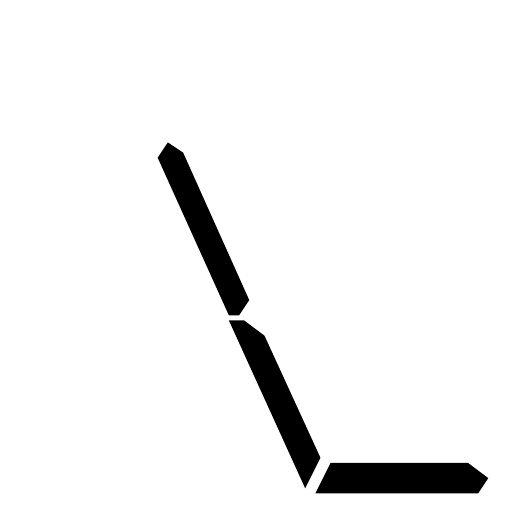} 
&
 \includegraphics[height=0.055\textwidth,width=0.055\textwidth]{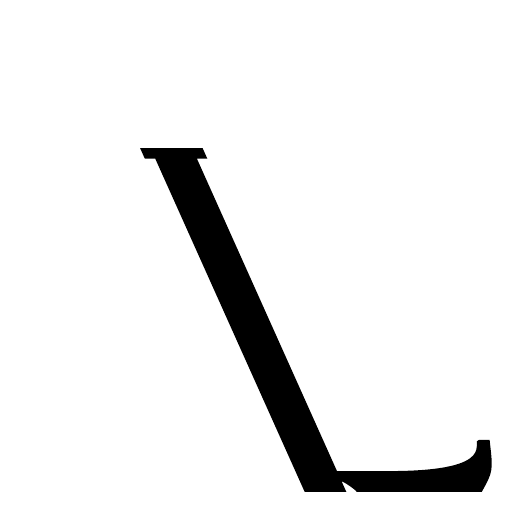}  
 &
 \includegraphics[height=0.055\textwidth,width=0.055\textwidth]{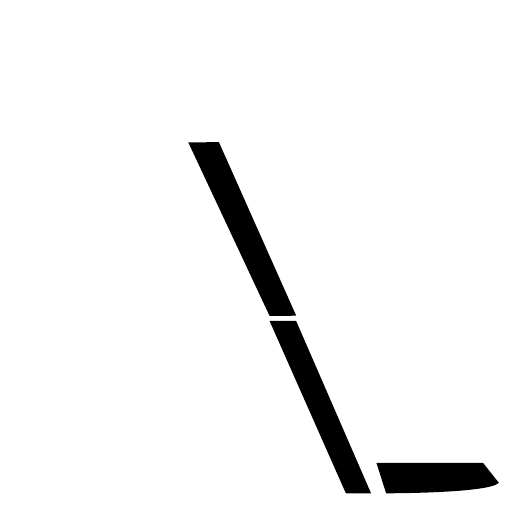} 
 &
 \includegraphics[height=0.055\textwidth,width=0.055\textwidth]{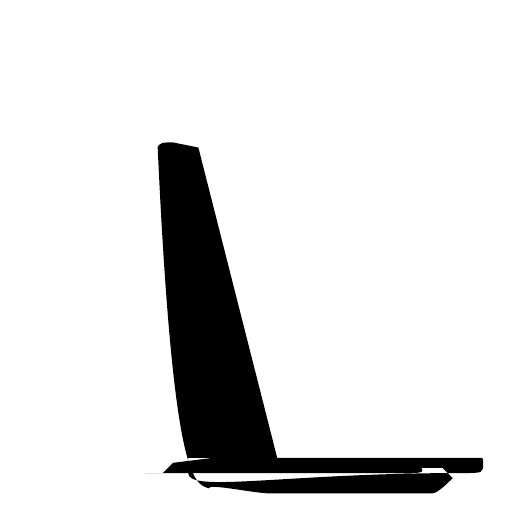} \\
\includegraphics[height=0.055\textwidth,width=0.055\textwidth]{images/few_shot_gen/CVPR/pdfs/4483_ref0.pdf} 
&
 \includegraphics[height=0.055\textwidth,width=0.055\textwidth]{images/few_shot_gen/CVPR/pdfs/4483_ref1.pdf} 
&
%  \includegraphics[height=0.055\textwidth,width=0.055\textwidth]{images/few_shot_gen/CVPR/pdfs/4483_ref2.pdf} 
% &
%  \includegraphics[height=0.055\textwidth,width=0.055\textwidth]{images/few_shot_gen/CVPR/pdfs/4483_ref3.pdf} 
% &
 \includegraphics[height=0.055\textwidth,width=0.055\textwidth]{images/few_shot_gen/CVPR/pdfs/4483_gt.pdf} 
&
 \includegraphics[height=0.055\textwidth,width=0.055\textwidth]{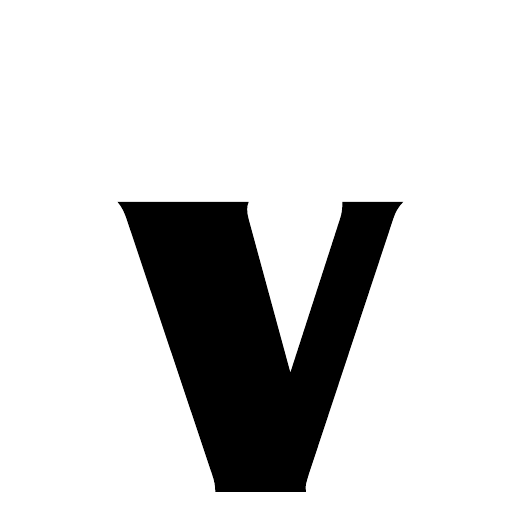} 
 &
 \includegraphics[height=0.055\textwidth,width=0.055\textwidth]{images/few_shot_gen/CVPR/pdfs/4483_ours.pdf} 
 &
 \includegraphics[height=0.055\textwidth,width=0.055\textwidth]{images/few_shot_gen/CVPR/pdfs/4483_dvfv2.pdf} \\
\includegraphics[height=0.055\textwidth,width=0.055\textwidth]{images/few_shot_gen/CVPR/pdfs/2546_ref0.pdf} 
&
 \includegraphics[height=0.055\textwidth,width=0.055\textwidth]{images/few_shot_gen/CVPR/pdfs/2546_ref1.pdf} 
&
%  \includegraphics[height=0.055\textwidth,width=0.055\textwidth]{images/few_shot_gen/CVPR/pdfs/2546_ref2.pdf} 
% &
%  \includegraphics[height=0.055\textwidth,width=0.055\textwidth]{images/few_shot_gen/CVPR/pdfs/2546_ref3.pdf} 
% &
 \includegraphics[height=0.055\textwidth,width=0.055\textwidth]{images/few_shot_gen/CVPR/pdfs/2546_gt.pdf} 
&
 \includegraphics[height=0.055\textwidth,width=0.055\textwidth]{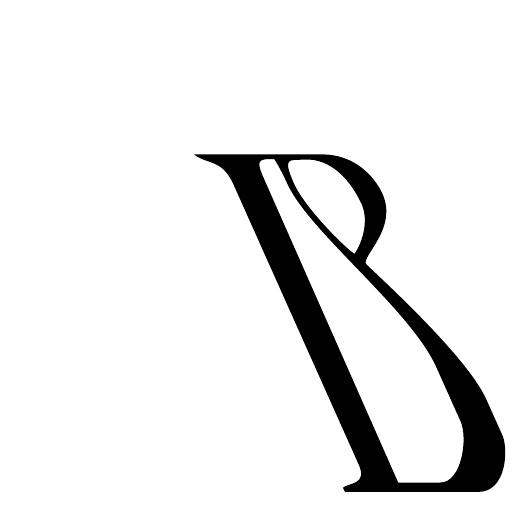}  
 &
 \includegraphics[height=0.055\textwidth,width=0.055\textwidth]{images/few_shot_gen/CVPR/pdfs/2546_ours.pdf} 
 &
 \includegraphics[height=0.055\textwidth,width=0.055\textwidth]{images/few_shot_gen/CVPR/pdfs/2546_dvfv2.pdf} 
 \\
 \includegraphics[height=0.055\textwidth,width=0.055\textwidth]{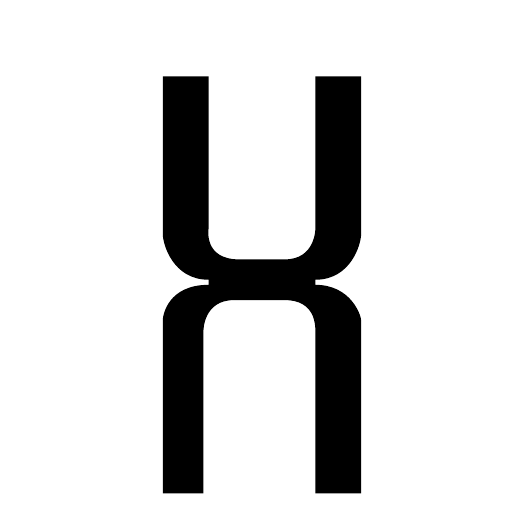} 
&
 \includegraphics[height=0.055\textwidth,width=0.055\textwidth]{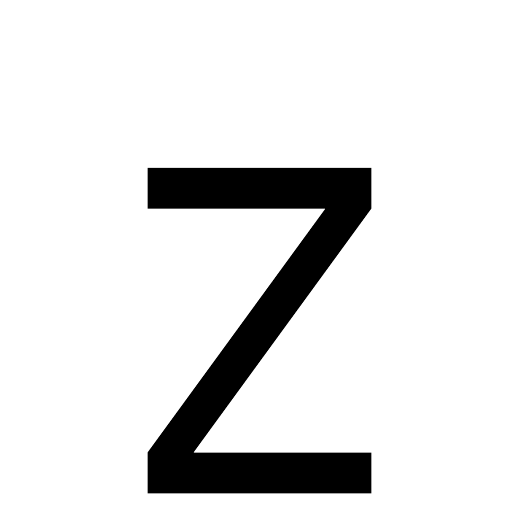} 
&
%  \includegraphics[height=0.055\textwidth,width=0.055\textwidth]{images/few_shot_gen/CVPR/pdfs/70_ref2.pdf} 
% &
%  \includegraphics[height=0.055\textwidth,width=0.055\textwidth]{images/few_shot_gen/CVPR/pdfs/70_ref3.pdf} 
% &
 \includegraphics[height=0.055\textwidth,width=0.055\textwidth]{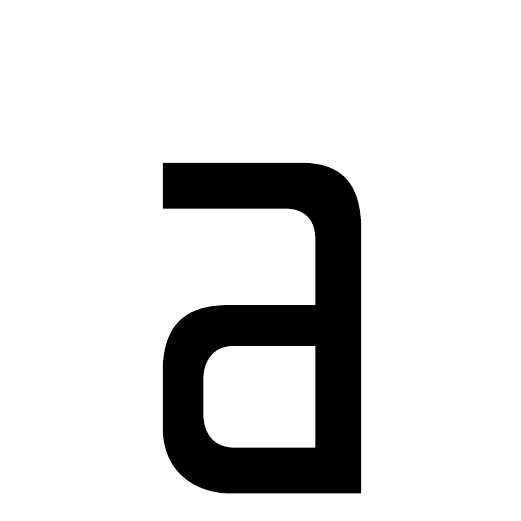} 
&
 \includegraphics[height=0.055\textwidth,width=0.055\textwidth]{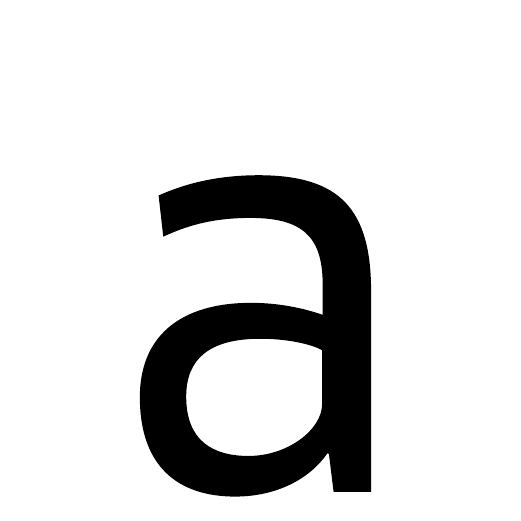}  
 &
 \includegraphics[height=0.055\textwidth,width=0.055\textwidth]{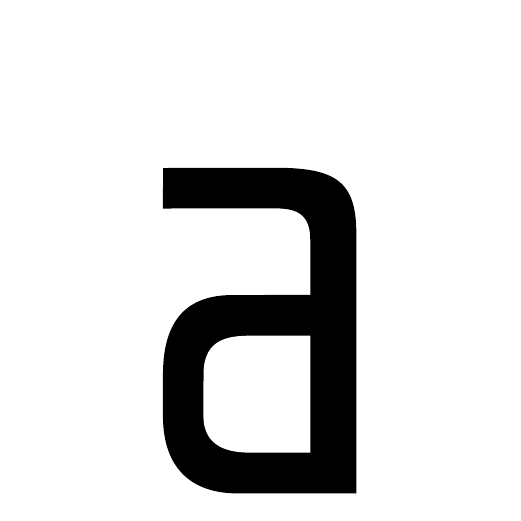} 
 &
 \includegraphics[height=0.055\textwidth,width=0.055\textwidth]{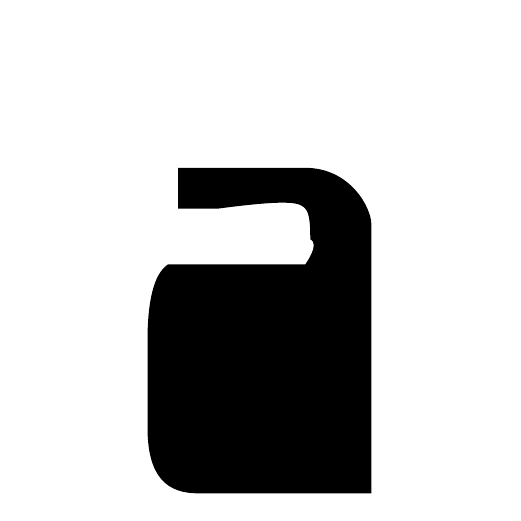} \\

 \includegraphics[height=0.05\textwidth,width=0.05\textwidth]{images/few_shot_gen/CVPR/pdfs/4343_ref0.pdf} 
&
 \includegraphics[height=0.05\textwidth,width=0.05\textwidth]{images/few_shot_gen/CVPR/pdfs/4343_ref1.pdf} 
&
%  \includegraphics[height=0.05\textwidth,width=0.05\textwidth]{images/few_shot_gen/CVPR/pdfs/4343_ref2.pdf} 
% &
%  \includegraphics[height=0.05\textwidth,width=0.05\textwidth]{images/few_shot_gen/CVPR/pdfs/4343_ref3.pdf} 
% &
 \includegraphics[height=0.05\textwidth,width=0.05\textwidth]{images/few_shot_gen/CVPR/pdfs/4343_gt.pdf} 
&
 \includegraphics[height=0.05\textwidth,width=0.05\textwidth]{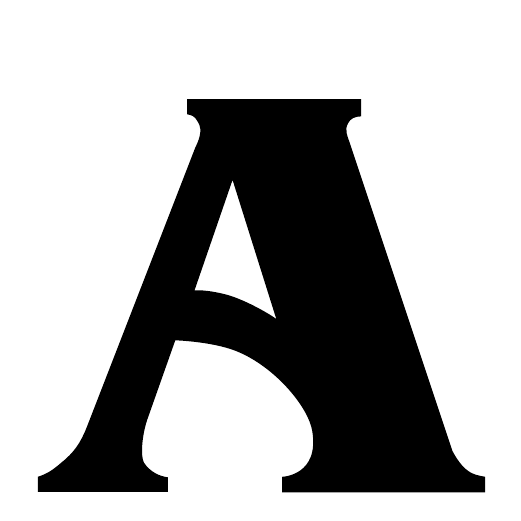}   
 &
 \includegraphics[height=0.05\textwidth,width=0.05\textwidth]{images/few_shot_gen/CVPR/pdfs/4343_ours.pdf} 
 &
 \includegraphics[height=0.05\textwidth,width=0.05\textwidth]{images/few_shot_gen/CVPR/pdfs/4343_dvfv2.pdf} \\
\includegraphics[height=0.06\textwidth,width=0.06\textwidth]{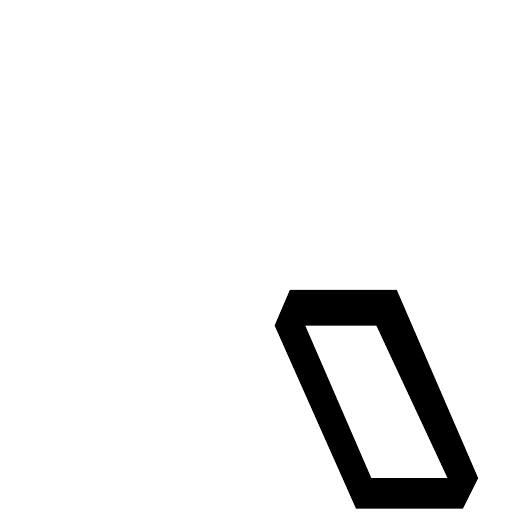} 
&
 \includegraphics[height=0.05\textwidth,width=0.05\textwidth]{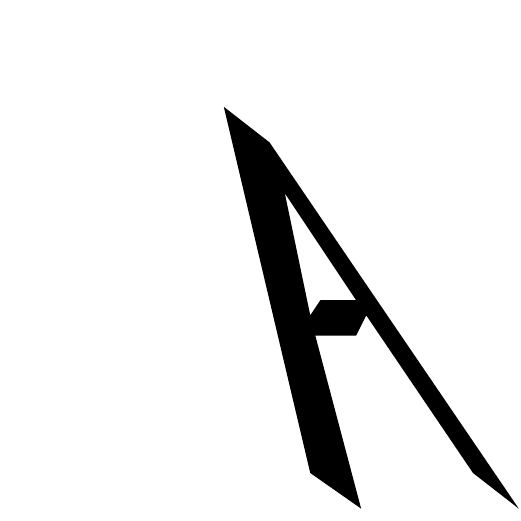} 
&
%  \includegraphics[height=0.06\textwidth,width=0.06\textwidth]{images/few_shot_gen/CVPR/pdfs/2215_ref2.pdf} 
% &
%  \includegraphics[height=0.06\textwidth,width=0.06\textwidth]{images/few_shot_gen/CVPR/pdfs/2215_ref3.pdf} 
% &
 \includegraphics[height=0.06\textwidth,width=0.06\textwidth]{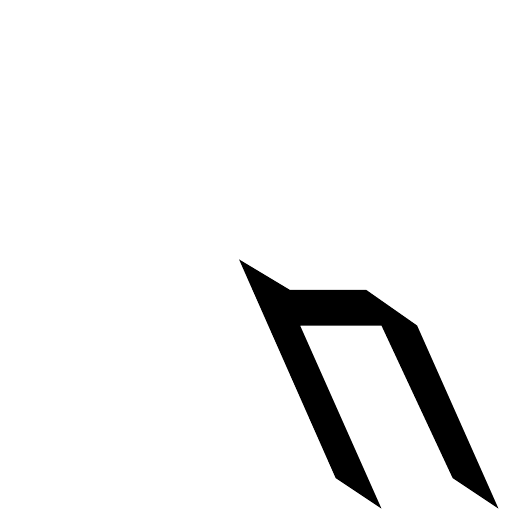} 
&
 \includegraphics[height=0.06\textwidth,width=0.06\textwidth]{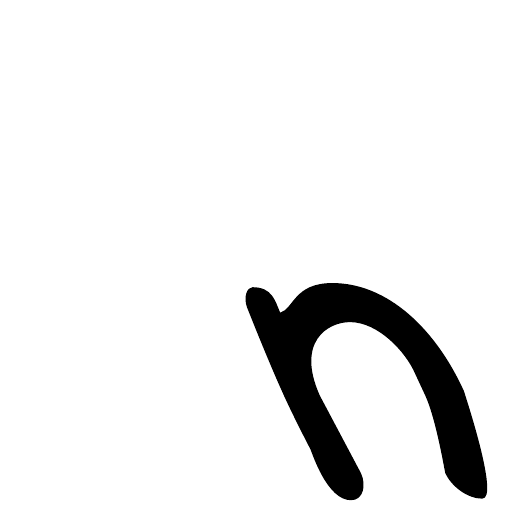} 
 &
 \includegraphics[height=0.06\textwidth,width=0.06\textwidth]{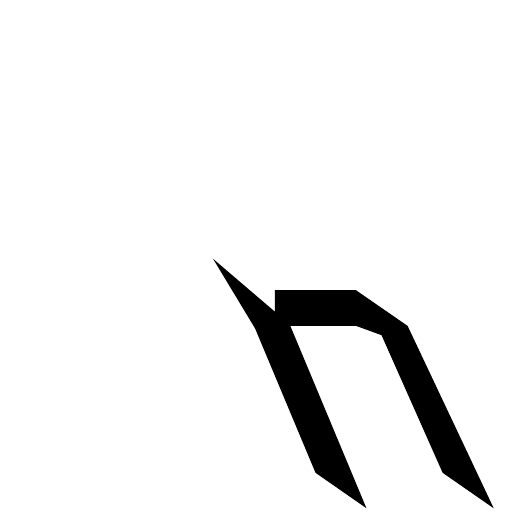} 
 &
 \includegraphics[height=0.06\textwidth,width=0.06\textwidth]{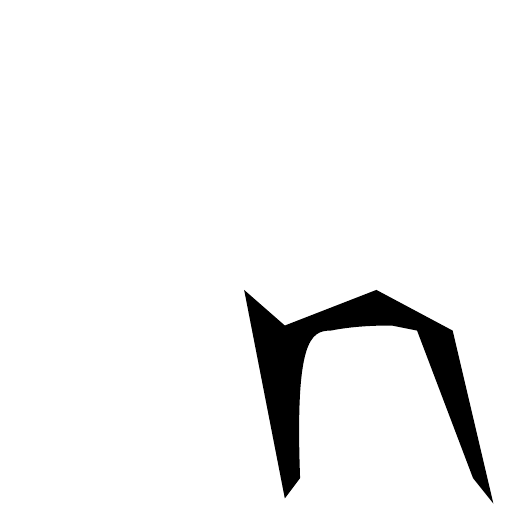} \\
 \includegraphics[height=0.055\textwidth,width=0.055\textwidth]{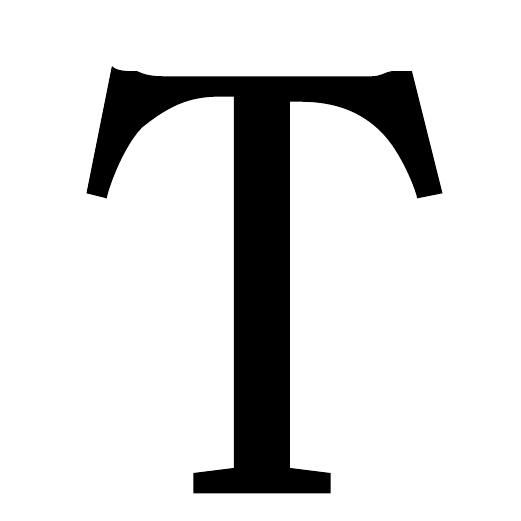} 
&
 \includegraphics[height=0.055\textwidth,width=0.055\textwidth]{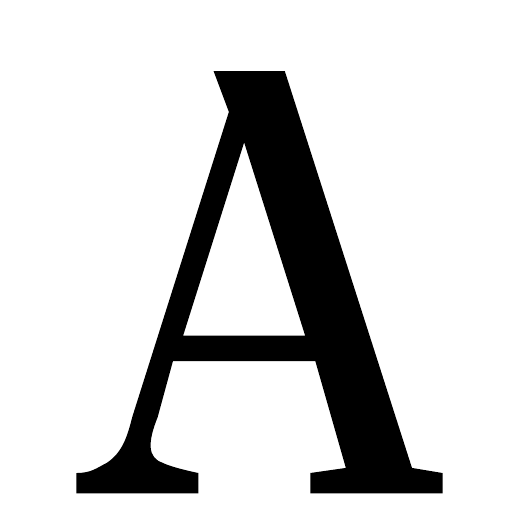} 
&
%  \includegraphics[height=0.055\textwidth,width=0.055\textwidth]{images/few_shot_gen/CVPR/pdfs/8_ref2.pdf} 
% &
%  \includegraphics[height=0.055\textwidth,width=0.055\textwidth]{images/few_shot_gen/CVPR/pdfs/8_ref3.pdf} 
% &
 \includegraphics[height=0.055\textwidth,width=0.055\textwidth]{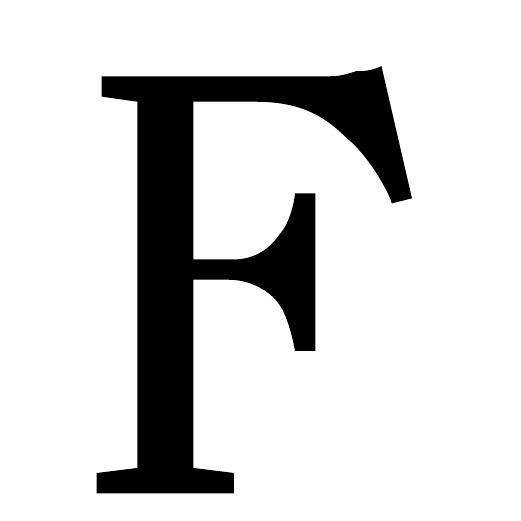} 
&
 \includegraphics[height=0.055\textwidth,width=0.055\textwidth]{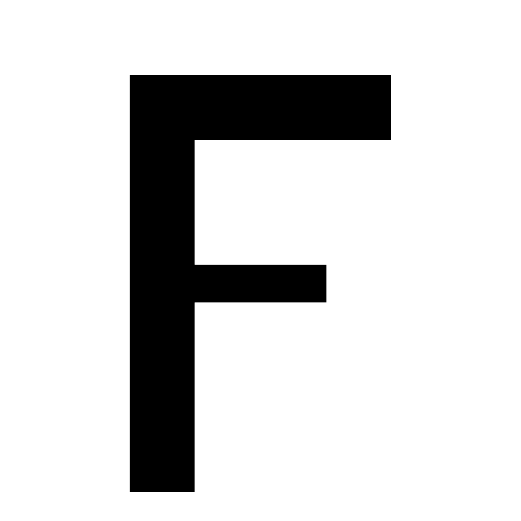}
 &
 \includegraphics[height=0.055\textwidth,width=0.055\textwidth]{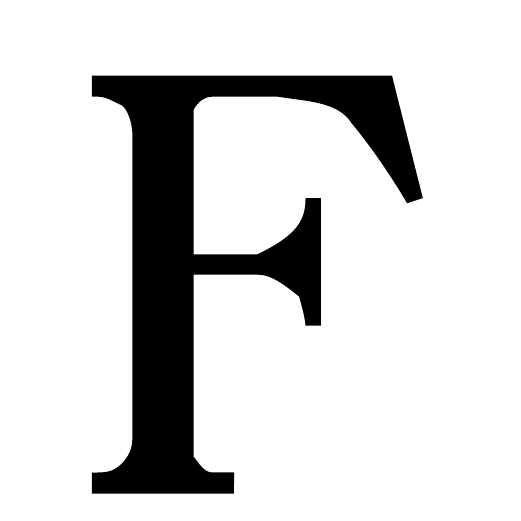} 
 &
 \includegraphics[height=0.055\textwidth,width=0.055\textwidth]{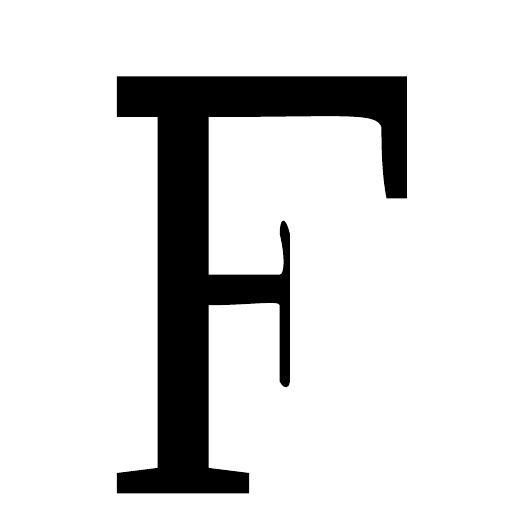} \\
 \includegraphics[height=0.055\textwidth,width=0.055\textwidth]{images/few_shot_gen/CVPR/pdfs/4842_ref0.pdf} 
&
 \includegraphics[height=0.055\textwidth,width=0.055\textwidth]{images/few_shot_gen/CVPR/pdfs/4842_ref1.pdf} 
&
%  \includegraphics[height=0.055\textwidth,width=0.055\textwidth]{images/few_shot_gen/CVPR/pdfs/4842_ref2.pdf} 
% &
%  \includegraphics[height=0.055\textwidth,width=0.055\textwidth]{images/few_shot_gen/CVPR/pdfs/4842_ref3.pdf} 
% &
 \includegraphics[height=0.055\textwidth,width=0.055\textwidth]{images/few_shot_gen/CVPR/pdfs/4842_gt.pdf} 
&
 \includegraphics[height=0.055\textwidth,width=0.055\textwidth]{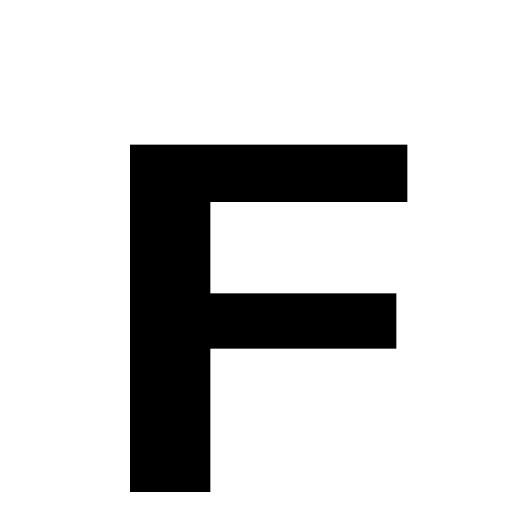}
 &
\includegraphics[height=0.055\textwidth,width=0.055\textwidth]{images/few_shot_gen/CVPR/pdfs/4842_ours.pdf} 
&
\includegraphics[height=0.055\textwidth,width=0.055\textwidth]{images/few_shot_gen/CVPR/pdfs/4842_dvfv2.pdf} \\
 \includegraphics[height=0.055\textwidth,width=0.055\textwidth]{images/few_shot_gen/CVPR/pdfs/4000_ref0.pdf} 
&
 \includegraphics[height=0.05\textwidth,width=0.05\textwidth]{images/few_shot_gen/CVPR/pdfs/4000_ref1.pdf} 
&
%  \includegraphics[height=0.055\textwidth,width=0.055\textwidth]{images/few_shot_gen/CVPR/pdfs/4000_ref2.pdf} 
% &
%  \includegraphics[height=0.055\textwidth,width=0.055\textwidth]{images/few_shot_gen/CVPR/pdfs/4000_ref3.pdf} 
% &
 \includegraphics[height=0.055\textwidth,width=0.055\textwidth]{images/few_shot_gen/CVPR/pdfs/4000_gt.pdf} 
&
 \includegraphics[height=0.055\textwidth,width=0.055\textwidth]{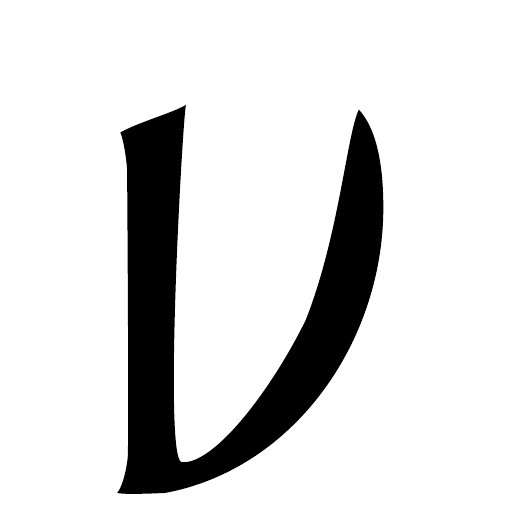}
 &
 \includegraphics[height=0.055\textwidth,width=0.055\textwidth]{images/few_shot_gen/CVPR/pdfs/4000_ours.pdf} 
 &
 \includegraphics[height=0.055\textwidth,width=0.055\textwidth]{images/few_shot_gen/CVPR/pdfs/4000_dvfv2.pdf} \\
    \end{tabular}
    \vspace{-1mm}
    \caption{Few-shot style transfer results. From left to right, we show the reference glyphs (2 out of 4) belonging to a novel font style, the artist-made (``ground-truth/ GT'') glyphs, the nearest-neighbours (``NNs'') to GT in the training data, our generated ones, and DeepVecFont-v2 (DVF-v2) \cite{wang2023deepvecfont} \label{fig:font_style_transfer_supp}}
    \vspace{-1em}
\end{figure}

\paragraph{Additional comparisons with DeepVecFont-v2 \cite{wang2023deepvecfont}.} 
Please see Figure \ref{fig:font_style_transfer_supp} for more comparisons with DeepVecFont-v2 \cite{wang2023deepvecfont} on the task of few-shot font style transfer.

\begin{figure}[t]
\centering
\includegraphics[width=0.97\linewidth]{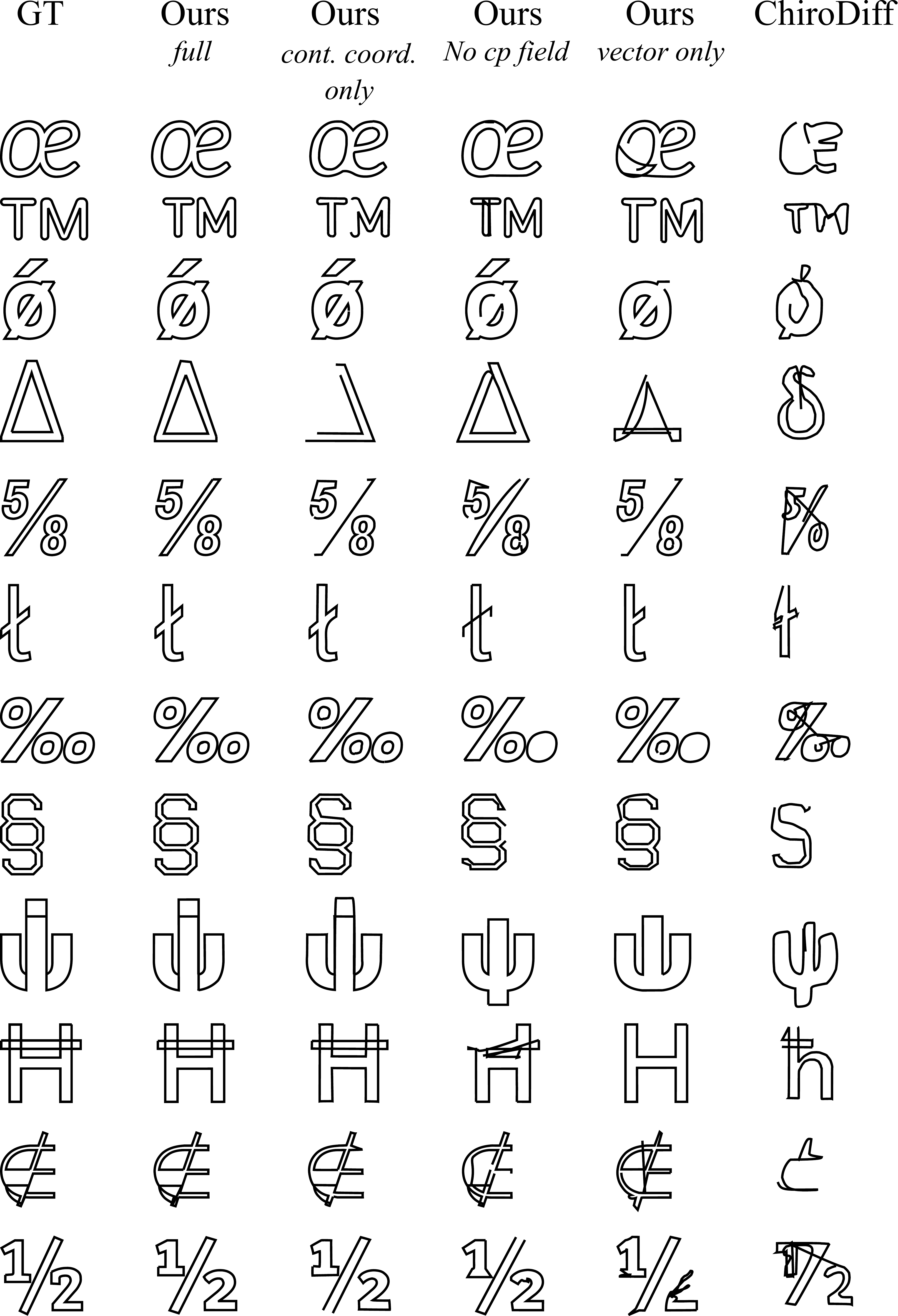}
\vspace{-2mm}
\caption{\rev{Glyph generation results for test cases from the Google font dataset. We compare our method to ChiroDiff \cite{das2023chirodiff} and degraded variants of our method. Our full method is able to generate glyphs that are much closer to artist-made (``ground-truth''/``GT'') ones compared to alternatives. }
}
\label{fig:missing_testset_comp_supp}
\vspace{-3mm}
\end{figure}

\paragraph{Additional comparisons with ChiroDiff \cite{das2023chirodiff}.} 
Please see Figure \ref{fig:missing_testset_comp_supp} for more comparisons with ChiroDiff \cite{das2023chirodiff} on the task of missing unicode generation.

\end{document}